\documentclass[letterpaper,12pt]{article}

\usepackage{amssymb, ipa, examples, chicago}

\newcommand{\ling}[1]{\textsl{#1}}
\newcommand{\C}[1]{{\sc #1}}

\newcommand{\OTempty}{$\square$} 

\newcommand{\undercirc}[1]{\oalign{#1\crcr
	\hidewidth\underring\hidewidth}}

\title{OT SIMPLE --\\ a construction-kit approach to Optimality Theory
implementation} 
\author{{\sc Markus Walther} \\ Heinrich-Heine-Universit\"{a}t D\"{u}sseldorf,
Germany\\ \texttt{walther@ling.uni-duesseldorf.de}}
\date{}
\begin{document}
%
\enlargethispage{4cm}
\thispagestyle{empty}
\begin{center}
\mbox{}\\[-35mm]
\huge {\bf Theorie des Lexikons} \\[6mm]
\mbox{Arbeiten~des~Sonderforschungsbereichs~282} \\[30mm]
\Large 
Nr. 88 \\[12mm]
{\bf OT SIMPLE -- \\[6mm]
 A construction-kit approach to Optimality Theory implementation} \\[12mm]
Markus Walther \\[30mm]
Oktober 1996 \\[17mm]
Projekt C6 \\[17mm]
Seminar f\"ur Allgemeine Sprachwissenschaft \\
Heinrich-Heine-Universit\"at D\"usseldorf \\
Universit\"atsstrasse 1 \\
40225 D\"usseldorf \\[4mm]
e-mail: walther@ling.uni-duesseldorf.de
\end{center}
\clearpage
\thispagestyle{empty}
\mbox{}
\clearpage
\pagenumbering{roman}
\maketitle
\begin{abstract}
This paper details a simple approach to the implementation of Optimality Theory
(OT, \citeNP{prince:smolensky:93}) on a computer, in part reusing standard system
software. In a nutshell, OT's GENerating source is implemented as a
BinProlog program 
interpreting a context-free specification of a GEN structural grammar
according to a user-supplied 
input form. The resulting set of textually flattened candidate tree
representations is passed to the CONstraint stage. Constraints are
implemented by finite-state transducers specified as `sed' stream
editor scripts that typically map ill-formed portions of the candidate
to violation marks. EVALuation of candidates reduces to simple sorting:
the violation-mark-annotated output leaving CON is fed into `sort', which
orders candidates on the basis of the violation vector column of each
line, thereby bringing the optimal candidate to the top. This approach
gave rise to OT SIMPLE, the first freely available software tool for the OT framework to provide
generic facilities for both GEN and CONstraint definition.   
Its practical applicability is demonstrated by modelling the OT
analysis of apparent subtractive pluralization in Upper Hessian
presented in \citeN{golston:wiese:95}.   
\end{abstract}
\newpage
\enlargethispage{2cm}
\thispagestyle{empty}
\tableofcontents
\newpage
\pagenumbering{arabic}
\section{Introduction}
This paper shows how to implement classical Optimality Theory (OT,
\citeNP{prince:smolensky:93}) in a particularly simple fashion
on a computer.%
\footnote{The research described in this paper has been carried out in
the {\em Sonderforschungsbereich 282
``Theorie des Lexikons'', Projekt C6 Prosodische Morphologie}, funded by the
German Research Agency (DFG). The author is grateful to Petra Barg,
James Kilbury, Thomas Klein and Richard Wiese, both for useful comments and
a thorough reading of the manuscript. The usual disclaimers apply. The
latest version of the OT SIMPLE software is available
on the Internet
(\texttt{http://www.phil-fak.uni-duesseldorf.de/$\sim$walther/otsimple.html}).}

 OT is an important emergent theoretical paradigm in
linguistics that by now underlies much research in both phonology and
morphology, while already being assumed in some work in syntax as
well.%
\footnote{See the current OT bibliography in the electronic archive at
\texttt{http://ruccs.rutgers.edu/roa.html} for a comprehensive
list of papers.}
For convenience, we nevertheless provide a quick overview
of the key features of this new theoretical paradigm in what follows,
with an eye towards its technical side.
\subsection{Optimality Theory in a nutshell}
OT proposes a general framework for the
theoretical analysis of natural language phenomena. 
In OT's conceptual model of natural language computation, an underlying
input form is first structurally enriched by a distinguished component
called GEN, yielding a possibly infinite set of structured output
candidates. OT per se is not committed to specific assumptions regarding
substantive aspects of the range and nature of the structure assigned
to inputs, but one of the requirements for GEN formulated so far,
`Freedom of Analysis' \cite{mccarthy:prince:93} in effect demands
that it should be sufficiently inclusive and rich. 

Candidates are then judged by an ensemble of 
universal wellformedness constraints, named the CON set in OT
terminology. Judgement consists of pairing each candidate with a
violation vector that records the degree to which the candidate
deviates from the wellformedness perspective 
imposed by each constraint. As a graphical rendering of degree of
violation, OT usually employs strings of asterisks, each asterisk or
violation mark indicating a constraint-specific unit of violation.
Depending on the nature of the constraint in question, such violation
may be {\em binary}, in that substructures of a candidate are only ever classified as obeying
or not obeying the constraint, emitting either no violation star or a
single one per substructure. An example of this type of all-or-nothing constraint
would be \C{Ons}, which states that syllables must have onsets \cite[16]{prince:smolensky:93}.
Alternatively, constraints may show {\em gradient} 
behaviour, typically using some abstract distance metric to calculate the deviation
from an ideal, e.g. that edges of two constituents such as affix and
word must coincide perfectly (cf. \C{Edgemost}, \citeNP[p.~35]{prince:smolensky:93}). These constraint types emit a number of violation marks that is proportional to the
actual distance found when assessing a candidate.%
\footnote{Actually, as \citeN{ellison:94} shows, some constraints including
\C{Edgemost}, that are described as being gradient, can be substituted
by binary equivalents, but others, e.g. the members of the \C{Align}
constraint family cannot (cf. \citeNP{ellison:95}). Another
issue to be kept separate from the question of gradienthood is that binary constraints 
will still result in multiple violation marks per entire candidate if
the substructure targeted by the constraint has multiple occurrences
in the candidates (e.g. polysyllabic words for \C{Ons}).} 

According to OT, language-specific grammars differ from each other in
terms of the ranking of the constraint set. The purpose of ranking, an ordering
relation over CON, is to reflect 
the relative importance of a constraint's judgement in situations
where judgements conflict with each other. As OT's strict domination
regime prescribes, a candidate violating a higher-ranked 
constraint to a lesser degree wins over all competing candidates
which incur more violation marks. Strict domination entails that
the competing candidates in this scenario cannot evade their looser
status by any amount of `compensatory' non-violation w.r.t. lower-ranked
constraints. 
However, in a situation where several candidates tie in
terms of incurring the same degree of current top-most constraint violation,
the entire set of those candidates is passed down to the 
next constraint in the hierarchy for further evaluation in the same vein,
and so on recursively. Since the constraint hierarchy is assumed to be
finite, the whole process is bound to terminate, in the worst case
using the lowest-ranked constraint to discern winning candidates. The
outcome of this evaluation procedure over violation vectors associated
with candidates, EVAL,%
\footnote{Note a slight difference in terminological meaning: 
While \citeN{prince:smolensky:93} use CON to denote the set of
ranked constraints, in this paper the term is broadened to include the
{\em action} of constraints on candidates to formally determine
constraint violations. Originally this part was delegated to
EVAL, which in the present 
setting is conversely limited to the evaluation of the violation patterns for purposes of
finding the optimal candidates. The main reason for this change of
meaning is that it better reflects the internal subdivision of OT
SIMPLE in terms of technical components.}
then is a set of optimal output candidates for
each input which minimally violate all constraints relative to the 
given ranking. This set of optimal candidates will often, but not
necessarily, consist of a single element, the winner.
\subsection{Motivation and previous work}
While the amount of work in theoretical linguistics, in particular
in the subfield of phonology, that is based on 
OT's conceptualization of grammar has virtually exploded over the last
three years, it is surprising to see so little practical computational
work that tries to formally capture GEN and CON specification on
the one hand and simulate the computation of optimal candidates given
such specifications on the other hand. This
state of affairs probably explains in part why it is
that virtually no OT work passes the formal adequacy criterion of
\citeN{bird:91}, which demands that the description language used to
capture generalizations, i.e. constraints, should possess a formal syntax and
semantics in the strict, mathematical sense. If it were
easy to specify constraints formally and check their action on GEN
output by computer, the stronger claims one could then make with
respect to correctness proofs and guaranteed empirical coverage would
surely be an attractive selling point in anybody's analysis.
Additionally, such possibilities would seem to be very promising for
meaningful inter-theory comparison on the basis of a given phenomenon
or corpus, since, for example, analyses couched in the
inviolable constraint-based Declarative Phonology paradigm (cf. \citeNP{scobbie:91},
\citeANP[to appear]{scobbie.et.al:96}) are also frequently computer-implemented. Note that
this task is different from a formal comparison of the frameworks
themselves. See \citeN{ellison:94a} for just such a comparison of
Underspecification Theory and Optimality Theory, both default-based
frameworks, and default-free Exception Theory, bringing out surprising
similarities between the former two approaches as one of the results.

As a matter of fact, \citeN{ellison:94} published the first algorithm for
computing with OT, assuming regular candidate sets, the most
restrictive class in the well-known Chomsky hierarchy of formal
languages as the formal specification language for describing
both GEN(Input) and each constraint. The alphabet over which these regular sets
are defined are pieces of linguistic information such as segments,
syllable roles etc. paired with one of two numerical marks. The marks
indicate local violation (1) or local wellformedness (0). Since
the corresponding finite-state automata (FSAs), standard models
of regular sets, are just special, numerically weighted graphs,
constraint interaction corresponds to regular set intersection modulo
a rank-sensitive concatenation of the mark vectors of two compatible
graph arcs, which makes the augmented graph/FSA product operation
corresponding to intersection asymmetric.
A variant of a graph algorithm for solving the so-called 
`shortest paths problem' for weighted graphs then forms the core of
Ellison's proposal for EVAL. In brief, the term `shortest paths' here
denotes those graph arc traversals from a FSA start state to a final
state which minimize the sum of violation mark vectors accumulated on the
way. 

While Ellison's work filled an important gap in the original conception
of OT by providing the first proposal for a formal
specification language for constraints and giving algorithms for {\em
all} aspects of an OT computation from GEN input to winner set
determination, an implemented and practically usable software tool
that builds on Ellison's work is yet to become publicly available.  
\footnote{However, Mark Ellison (p.c.) has done a prototype
implementation by way of extending his as yet unpublished Typed
Regular Description Language (TRDL) formalism. He reports that there
is also work in progress at the Universities of Bielefeld and
Edinburgh on implementing variants of the original proposal.}

Another strand of work on the problem of computing with OT is
represented by \citeN[1995, 1996]{tesar:94}.
In his approach, the space of possible GEN structures is encoded as a
regular or context-free position structure grammar. GEN is formalized
as a set of matchings between an ordered string of input segments and
the terminals of the grammar, preserving the linear order of the
input. A dynamic programming algorithm is used to boost efficiency in
computing the optimal candidate, using the structure grammar to propose
partial structures with and without faithfulness violations and
assessing them via the ranked constraint set.  

One important difference, when compared to Ellison's approach, is that
Tesar's algorithms deal with (partial) candidates on the {\em object} level,
albeit in a somewhat more intelligent fashion than simple enumeration,
whereas Ellison is able to directly compute with an intensional
{\em description} of the entire candidate set, subject to the weak
restriction that it is formally regular. In general, an 
intensional approach seems more elegant and uniform, in this case employing
regular sets throughout all stages of Ellison's remarkably simple
algorithms, while it also has the theoretical advantage of halting
even if the winner set of optimal candidates should turn out to be
infinite. Faced  with this situation any complete enumerating approach
would not terminate, since infinite sets are not representable on the
object level. 
Still, for the usual situation of a finite winner set the dynamic
programming approach Tesar adopts to compute constraint action
ensures reasonably efficient algorithmic complexity. However, from the
presentation in \citeN[1996]{tesar:95} it is not entirely clear
how non-singleton winner sets are actually handled and what
restrictions w.r.t. cardinality may have to be assumed. More importantly
for our purposes, Tesar is not explicit on 
what formal description language to adopt for specifying constraints.
From severe locality requirements that his approach, unlike Ellison's,
must impose on constraints,%
\footnote{\citeN{tesar:96} states that for formally regular position
structure grammars, a constraint must be locally evaluable on the
basis of two consecutive positions in the linear position structure.
For context-free position structure grammars, locality amounts to
the requirement that at most one of the local tree configurations
$nonterminal\_mother[nonterm._1 \dots nonterm._N]$ or 
$nonterminal\_mother[terminal]$ may form the basis for deciding
constraint action.}  
it appears it should be some subset 
of the regular languages. Again, a practical and 
generally available software tool for specifying GEN and CON as well as
computing a winner set via EVAL is not known to exist. 

In contrast to the aforementioned general approaches to OT
computation, the two implemented tools of \citeN{andrews:95} and
\citeN{hammond:95} which {\em are} available focus on quite specific
narrow-scale applications. While Andrews focuses on two demonstration versions of
syllable theory and an application to Lardil (implementing parts of
\citeNP[ch.6-8]{prince:smolensky:93}),  Hammond is interested in the
contrastive behaviour of an OT-based syllabification strategy with
rankings for English vs. French. Neither comes with a concomitant
formal proposal for a sufficiently general constraint description language. 
Rather, GEN and the constraints of interest have been hard-wired 
somewhat ad hoc into the respective program code. While this is certainly a
viable move for the stated purposes, one cannot expect a
principled solution for the problem of general OT specification and
computation to be derivable from these works.

Finally, note that a software tool developed by
\citeN{raymond:hogan:96}, despite its name {\em Optimality Theory 
Interpreter}, does not provide any means to implement GEN, constraints
and constraint action -- the task under focus here -- restricting itself to the
comparatively easier task of mechanizing EVAL. Its use is
therefore limited to helping explore the consequences of reranking
etc. given {\em human} input for GEN's output candidates and violation
mark vectors. 
\subsection{Objectives}
With this quick survey of related work in place, let us now turn to
describe the major goals of OT SIMPLE, our software tool for practical
OT computation. The stated objectives that should
be met were defined as follows: OT SIMPLE should
\begin{itemize}
\item {\bf facilitate rigid specification and testing of truly formal
OT analyses}. The standard practice of graphical or natural language
specification of constraints paired with their intuitive evaluation
against a few hand-selected candidates is both error-prone and
bound to fail over larger corpora and constraint sets. Put
differently, human theorists on the one hand  are far too smart to bother
themselves with all the crazy structural possibilities in the
GEN structural space or to consider every dull technical detail of every
constraint during constraint evaluation in a mechanical fashion, on
the other hand they are too limited resources to reliably check the
zillions of candidates against a full constraint hierarchy. 
\item {\bf be simple to build and reuse existing pieces of standard
software}. The original OT description is intuitively simple to grasp,
so a direct reflection of this simplicity in an implementation would
be favourable. Reusing existing standard software minimizes coding
effort and errors and helps attain greater portability to other
platforms.
\item {\bf stay close to familiar operational model and conceptual
entities}. Intuitions of the working linguist on how OT works are
presumably shaped by the original presentation in
\citeN{prince:smolensky:93} more than by anything else, so a user
interface which overtly reflects 
these intuitions to a reasonably high degree should help alleviate the barrier
to strict formalization and also encourage using OT SIMPLE for
pedagogical purposes. It certainly would be user-friendly to see GEN
actually feed into a constraint hierarchy on the screen, which assigns
visible asterisks and in turn feeds into EVAL, with constituent
structure displays showing the winner.
\item {\bf be sufficiently general}. One would like to be free to
specify a wide range of different constraint types and have sufficient
flexibility for defining the structure of GEN in order to suit the
demands of a concrete analysis.
\item {\bf make reranking and inspection of intermediate stages in
 constraint evaluation easy}. This is a definite must for developing
formal, implemented analyses incrementally, testing and debugging
constraints as the theory develops. One should be able to freely
rerank constraints and quickly see the effects in order to be able to
experiment with factorial typologies and alternative analyses.
\end{itemize}
The next section shows in more detail to what extent these demands
were met in constructing OT SIMPLE.
\newpage
\section{General architecture}
Here is an overview of OT SIMPLE.
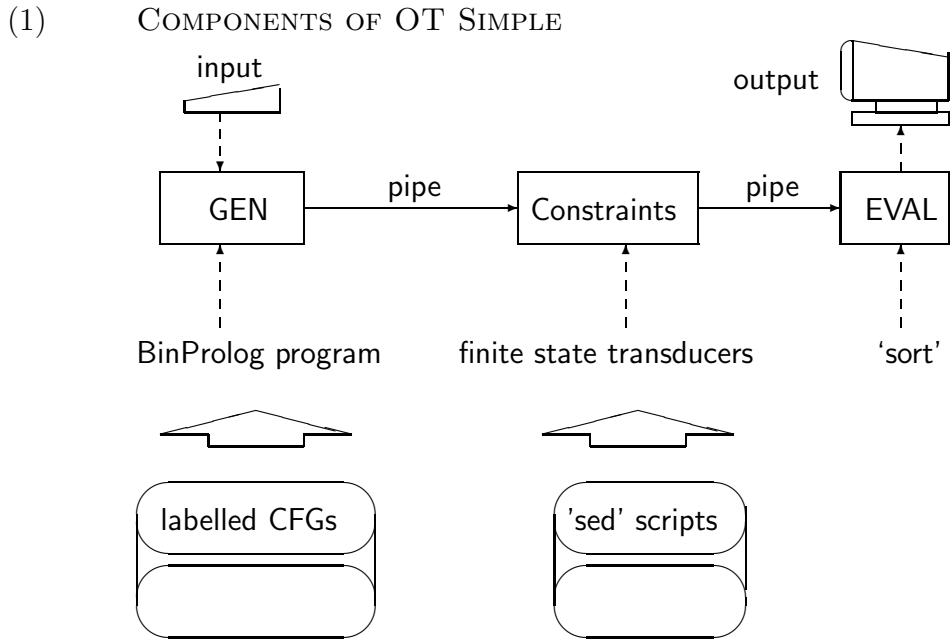
\begin{examples}
\item \label{OTSimpleOverview}
{\sc Components of OT Simple} \\
\setlength{\unitlength}{0.012500in}%
\begingroup\makeatletter\ifx\SetFigFont\undefined
\def\x#1#2#3#4#5#6#7\relax{\def\x{#1#2#3#4#5#6}}%
\expandafter\x\fmtname xxxxxx\relax \def\y{splain}%
\ifx\x\y   
\gdef\SetFigFont#1#2#3{%
  \ifnum #1<17\tiny\else \ifnum #1<20\small\else
  \ifnum #1<24\normalsize\else \ifnum #1<29\large\else
  \ifnum #1<34\Large\else \ifnum #1<41\LARGE\else
     \huge\fi\fi\fi\fi\fi\fi
  \csname #3\endcsname}%
\else
\gdef\SetFigFont#1#2#3{\begingroup
  \count@#1\relax \ifnum 25<\count@\count@25\fi
  \def\x{\endgroup\@setsize\SetFigFont{#2pt}}%
  \expandafter\x
    \csname \romannumeral\the\count@ pt\expandafter\endcsname
    \csname @\romannumeral\the\count@ pt\endcsname
  \csname #3\endcsname}%
\fi
\fi\endgroup
\begin{picture}(341,250)(5,585)
\thinlines
\put(305,823){\oval( 10, 24)[tl]}
\put(305,823){\oval( 10, 26)[bl]}
\put(305,800){\framebox(40,5){}}
\put(315,805){\framebox(25,5){}}
\put(345,830){\line( 0,-1){ 20}}
\put(345,810){\line(-1, 0){ 40}}
\put(305,810){\line( 0, 1){ 25}}
\put(305,835){\line( 6,-1){ 39.730}}
\put(345,830){\line( 0, 1){  0}}
\put(255,815){\makebox(0,0)[lb]{\smash{\SetFigFont{12}{14.4}{sf}output}}}
\put(220,635){\oval(80,30)}
\put(220,600){\oval(80,30)}
\put( 55,635){\oval(100,30)}
\put( 55,600){\oval(100,30)}
\multiput( 40,805)(0.00000,-7.14286){4}{\line( 0,-1){  3.571}}
\put( 40,780){\vector( 0,-1){0}}
\put( 15,750){\framebox(60,30){}}
\put(165,750){\framebox(75,30){}}
\put(300,750){\framebox(45,30){}}
\put( 75,765){\vector( 1, 0){ 90}}
\put(240,765){\vector( 1, 0){ 60}}
\multiput( 40,715)(0.00000,7.77778){5}{\line( 0, 1){  3.889}}
\put( 40,750){\vector( 0, 1){0}}
\multiput(210,715)(0.00000,7.77778){5}{\line( 0, 1){  3.889}}
\put(210,750){\vector( 0, 1){0}}
\multiput(325,715)(0.00000,7.77778){5}{\line( 0, 1){  3.889}}
\put(325,750){\vector( 0, 1){0}}
\put( 35,665){\line( 0, 1){  5}}
\put( 35,670){\line(-1, 0){ 20}}
\put( 15,670){\line( 4, 1){ 40}}
\put( 55,680){\line( 4,-1){ 40}}
\put( 95,670){\line(-1, 0){ 20}}
\put( 75,670){\line( 0,-1){  5}}
\put( 75,665){\line(-1, 0){ 40}}
\put(195,665){\line( 0, 1){  5}}
\put(195,670){\line(-1, 0){ 20}}
\put(175,670){\line( 4, 1){ 40}}
\put(215,680){\line( 4,-1){ 40}}
\put(255,670){\line(-1, 0){ 20}}
\put(235,670){\line( 0,-1){  5}}
\put(235,665){\line(-1, 0){ 40}}
\put(180,635){\line( 0,-1){ 35}}
\put(260,640){\line( 0,-1){ 35}}
\put(  5,635){\line( 0,-1){ 35}}
\put(105,635){\line( 0,-1){ 35}}
\put( 25,805){\line( 0, 1){  5}}
\put( 25,810){\line( 6, 1){ 39.730}}
\put( 65,815){\line( 0,-1){ 10}}
\put( 65,805){\line(-1, 0){ 25}}
\put( 40,805){\line(-1, 0){ 15}}
\multiput(325,780)(0.00000,8.00000){3}{\line( 0, 1){  4.000}}
\put(325,800){\vector( 0, 1){0}}
\put( 35,760){\makebox(0,0)[lb]{\smash{\SetFigFont{12}{14.4}{sf}GEN}}}
\put(  5,700){\makebox(0,0)[lb]{\smash{\SetFigFont{12}{14.4}{sf}BinProlog program}}}
\put(140,700){\makebox(0,0)[lb]{\smash{\SetFigFont{12}{14.4}{sf}finite state transducers}}}
\put(315,700){\makebox(0,0)[lb]{\smash{\SetFigFont{12}{14.4}{sf}`sort' }}}
\put(170,760){\makebox(0,0)[lb]{\smash{\SetFigFont{12}{14.4}{sf}Constraints}}}
\put(310,760){\makebox(0,0)[lb]{\smash{\SetFigFont{12}{14.4}{sf}EVAL}}}
\put(110,770){\makebox(0,0)[lb]{\smash{\SetFigFont{12}{14.4}{sf}pipe}}}
\put(260,770){\makebox(0,0)[lb]{\smash{\SetFigFont{12}{14.4}{sf}pipe}}}
\put(185,630){\makebox(0,0)[lb]{\smash{\SetFigFont{12}{14.4}{sf}'sed' scripts}}}
\put( 15,630){\makebox(0,0)[lb]{\smash{\SetFigFont{12}{14.4}{sf}labelled CFGs}}}
\put( 30,820){\makebox(0,0)[lb]{\smash{\SetFigFont{12}{14.4}{sf}input}}}
\end{picture}
\end{examples}
Starting from left to right, GEN is realized by a BinProlog
(\citeNP[1996]{tarau:92}) program. Its static input takes the form of a
variant of context-free grammars (CFGs), which describe the space of possible structures
that can be erected over an input. CFG production rules may be
labelled with mnemonic markers, to be referred to in the dynamic input
to GEN, depicted above the keyboard symbol. The candidate output of GEN is in
the form of tree structures, using a flattened format suitable for textual
line-based representation. Each line contains one tree. 

This multi-line candidate set data stream is literally fed into the next
stage by means of a {\em pipe}. Simplifying somewhat, pipes are
constructs provided by several computer operating systems (OSes), in
particular, UNIX derivates, to connect the standard output of a
program (which would otherwise go to the screen) to the standard input
of another program (which would otherwise come from the keyboard).

The next stage provides an operational model of CON. Its main
ingredient is a cascade of finite-state transducers (FSTs), one for
each constraint.%
\footnote{For an introduction to
finite-state automata in general, and the special case represented by
FSTs as well as the notions of regular sets and expressions
used at various stages in the description of OT SIMPLE below, see e.g.
\citeN{hopcroft:ullman:79}.}
These are specified and cascaded using a piece of
standard software called `sed', a fast stream editor found in
particular in UNIX OSes. Specifically, `sed' provides commands for
substituting parts of a line that match a given pattern by new
strings. Sequences of these commands are stored in `sed' scripts, the
materialization of constraints that forms the static input to CON.
Importantly, pattern descriptions can be defined using the full power
of regular expressions; this gives the needed flexibility for a wide
range of phonological context types, therefore constituting the
essential ingredient which motivated the choice of `sed'. Descriptions
would typically characterize some illformed structural configuration,
and new strings would typically contain asterisks in place of the
particular configuration.
After the dynamic input to CON, i.e. the multi-line candidate stream,
has been enriched by `sed' actions according to its
constraint scripts, the ensuing violation-annoted output stream is again
connected via a pipe to the final component, EVAL. 

EVAL is nothing else than familiar sorting.%
\footnote{This equivalence was also alluded to in \citeN[p.~19]{prince:smolensky:93}.}
 It uses another piece of
standard software, a sorting routine named `sort' that is again
available in all UNIX OSes and elsewhere.
Using `sort' exploits the fact that we can choose both the violation
symbol and its non-violation counterpart in such a way that
non-violation is ordered before violation in the standard ASCII
character encoding used on computers. Here, the
asterisk ($\ast$) is the violation character and its non-violation counterpart
is defined as the single quote character (\verb|'|).%
\footnote{With $\ast$ = ASCII code $42_{10}$ and \texttt{'} = ASCII
code $39_{10}$, the noted sorting order follows.}
Bringing the winner(s) to the top therefore amounts to simple textual
sorting of the lines of the annotated candidate stream according to
the violation vector column that comes last in each line. 
Since, in addition to the intrinsic sorting order between the two types
of marks, shorter column entries are sorted before longer ones
and individual constraint violation fields in the violation vector are always
formally separated by the non-violation mark \verb|'|, using sorting
as the device to effect minimal violation or optimization is
sound. The sorted output of `sort' can then e.g. be stored to a file,
viewed directly on the screen or fed into a graphical tree display
tool for convenience, with the first candidate line showing the
optimal output. 

After this tour-de-force through the general architecture of OT
SIMPLE, let us now proceed to describe how one can
fill both GEN and CON with substantive content, provide some input
and then see OT at work using this software tool.
\section{Specifying GEN}
GEN is the component of OT that is responsible for structural
enrichment of a bare input. The immediate question is therefore how to
specify the space of structural possibilities. In OT SIMPLE a context-free
phrase structure rule format is used for this purpose. This degree of
formal power should suffice even for some cases of syntax modelling. As is
well-known, context-free grammars (CFGs) can only describe tree
structures, which is therefore a limitation that OT SIMPLE currently
inherits (cf. \citeNP{tesar:96} for the same assumption). While more general
reentrant graph structures may be desirable for some multiply-linked structures
as used e.g. in autosegmental representations, we will see later that
their present omission is not significant for the important class of
automatic, i.e. predictable sharing or linking.
(\ref{SampleGenGrammar}) presents a particularly simple example, the
first part of a GEN grammar%
\footnote{This term is used interchangeably with `GEN structur(e/al)
grammar' in what follows. It is equivalent to \citeN{tesar:96}'s `position 
structure grammar'.}
 file \verb|hessian.gen|, which will be extended later. Note that the
\verb|.gen| file name extension is obligatory. We 
hasten to add that OT SIMPLE, like the overall OT {\em framework}
itself, is entirely neutral with respect to the substantive
constituent and rule inventory that a theorist may wish to adopt in
specifying the desired space of possible GEN structures. 
\begin{examples}
\item \label{SampleGenGrammar}
{\sc Sample GEN grammar} \\
\begin{verbatim}
startsymbol word.                           % 0
word ---> ft.                               % 1
word ---> ft, ft.                           % 2
                                            % 3
ft   ---> syl.                              % 4 
ft   ---> syl, syl.                         % 5
                                            % 6
syl  ---> m.                                % 7
syl  ---> 'Rt', m.                          % 8
syl  ---> 'Rt', m, m.                       % 9
syl  ---> 'Rt', m, m, 'Rt'.                 % 10
                                            % 11
m    ---> 'Rt'.                             % 12
                                            % 13
'Rt' ---> [].                               % 14
                                            % 15
a # 'Rt' ---> "SONORANT", "DORSAL".         % 16
t # 'Rt' ---> "SPREAD_GLOTTIS", "CORONAL".  % 17
\end{verbatim}
\end{examples}
The context-free productions in lines 1-5 define expansions of word,
foot (\verb|ft|) and
syllable (\verb|syl|) into their maximally binary lower constituents
in the familiar 
prosodic hierarchy, whereas subsyllabic structure  receives a moraic
treatment in lines 7-10, with a minimum of one mora (\verb|m|) and a
maximally bimoraic structure with one additional onset-position root
node (\verb|'Rt'|) and one extra coda position (line 10). Moras also dominate
root nodes (line 12). The first device  in this CFG format that is
special to OT appears in line 14. It specifies a production rule that
rewrites into a built-in \verb|[]| terminal symbol equivalent in
function to OT's empty structure symbol, \OTempty. Note that it is
under complete control of the GEN grammar writer which parts of the
structure should be eligible for free \OTempty insertion. Lines 16 and 
17, which specify alternative, non-empty expansions of root nodes
illustrate two more notational devices: first, terminal symbols like
the privative features \verb|SONORANT, DORSAL, SPREAD_GLOTTIS, CORONAL|
are distinguished from nonterminals by surrounding double quotes. More
importantly, arbitrary productions can be labelled with a symbol in
front of the hash sign (\verb|#|); thus, e.g. \verb|a| acts as a
shorthand reference label for the rule in line 16. Since rules in turn
are in familiar correspondence with local trees, labels can be
understood as triggering the presence of local trees in candidates.
Labels are the sole means to specify inputs. As we will see later in
concrete GEN usage, inputs in OT SIMPLE are nothing other than a
comma-separated list of such labels enclosed by square brackets, e.g.
\verb|[t,a]|.  

Any CFG by definition must specify a distinguished start symbol. If,
unlike line 0 in (\ref{SampleGenGrammar}), no overt \verb|startsymbol YourStartSymbol.|  
declaration is included, the startsymbol \verb|word| is assumed by
default. For those who are less familiar  with the Prolog naming and
file layout conventions that are also used in 
the GEN grammar file format, it may be worth emphasizing that each rule must be
terminated with a period symbol, that the arrow must consist of
exactly three consecutive minus signs followed by a greater-than
symbol, and that symbols with upper-case initials must be enclosed in
single quotes. Otherwise, the GEN grammar can be written in free format with
arbitrary numbers of whitespace characters between symbols, and
rules may appear in any order. Moreover, the rest of the line after a
percent sign is ignored and may contain comments.

In order to satisfy the `Freedom of Analysis' condition, a good GEN
structural grammar must be highly ambiguous. This property is
obtained in (\ref{SampleGenGrammar}), which allows a
large number of alternative structures to be built from terminal input seeds
such as \verb|[t,a]|. In (\ref{SampleGenGrammar}), all
categories except for \verb|m|(ora) have more than one expansion, and
with the help of free epenthesis quite baroque tree structures will be
realizable. One important additional OT concept that is not directly
visible in a GEN grammar, however, is {\em underparsing}. The reason for its
invisibility is that in OT SIMPLE underparsing is built-in: it automatically
pertains to all lefthand-side categories of labelled rules that can be
used as input, e.g. \verb|'Rt'| in (\ref{SampleGenGrammar}).  This
means that the set of different structural realizations w.r.t. the GEN
grammar will be multiplied with all possible underparsings of input
tokens used in the tree structures.  

Another point to be taken into account is the cardinality of the GEN
set in conjunction with the amount of freely insertible empty
structure. While not being the case for the rules in 
(\ref{SampleGenGrammar}), in OT SIMPLE GEN structural grammars are
generally allowed to contain recursive productions. If, in keeping
with the original presentation of OT, candidate structures are
actually enumerated one at a time, some means must be found to ensure
GEN termination. Especially with an infinite supply of empty symbols
GEN enumeration might otherwise never halt. Since for all practical
purposes a finite GEN set appears quite tolerable, OT SIMPLE
consciously opts for incompleteness here by treating a finite but
arbitrarily large number of empty symbols as an additional, but
invisible kind of technical input. This upper bound on the number of
epenthetic elements must be specified by the user. Hence the sum of
GEN's input proper and this technical input is also finite, and if the
GEN grammar contains no direct or indirect {\em left}-recursion,%
\footnote{The ban on left-recursion is due to the particularly simple
depth-first, left-to-right rule application strategy  which in turn 
straightforwardly follows from Prolog's internal proof strategy. Note that
right-recursive rules {\em are} admitted. However, we know
that every context-free language which does not include the empty
string can be described by a CFG without unit productions or
general left-recursion (a corollary of the Greibach normal form theorem),
hence in principle this situation is always avoidable. Because the
necessary grammar transformations may change parse trees, however, it
might be desirable to provide for alternative tree traversal
strategies directly. This is left for future improvements.}
the entire set of candidates will be
always finite. Finiteness of the GEN set also is a necessary precondition for
the implementation of EVAL in OT SIMPLE, which, we recall, is modelled by sorting.
While, as already noted, infinite GEN sets are not a problem for
Ellison's approach, the inspection of infinite set descriptions
themselves can be uninstructive. Individual candidates, in contrast,
are more easily understood; with them we firmly reside on the object
rather than on the description level. Since we know of no analyses which
crucially depend on infinite GEN sets, one need not be overly bothered
by this small restriction.

This concludes our discussion of GEN grammar specification. We will provide
specific details of how GEN is actually implemented in section
\ref{GenImplementation}, and show the concrete GEN user interface in section
\ref{GenInAction}. 
\section{Specifying constraints}
Constraints in OT SIMPLE act to annotate candidates
with violation vectors. Each column of the vector in turn
consists of a contiguous string of asterisks corresponding to actual
violations or a single quote for non-violation. Constraints operate on the
flattened tree format that is output by GEN for purposes of textual
representation, containing one candidate tree per line of GEN output.
While this is shown in more detail in section \ref{GenInAction}, here
are some typical candidates for immediate illustration:%
\footnote{Here and in the following, overlong candidate lines will be folded for
presentational purposes, as indicated by the backslash.}
\begin{examples}
\item \label{TypicalCandidates}
{\sc Typical Candidates}
\footnotesize
\begin{verbatim}
word(ft(syl(Rt(SPREAD_GLOTTIS,CORONAL),m(Rt(SONORANT,DORSAL))))).
word(ft(syl(Rt(SPREAD_GLOTTIS,CORONAL),m({Rt(SONORANT,DORSAL)})))).
word(ft(syl(Rt([]),m({Rt(SPREAD_GLOTTIS,CORONAL)}),\
    m({Rt(SONORANT,DORSAL)})))).
\end{verbatim}
\end{examples}
What matters for present purposes is the candidate format and not its content.
In this format a local tree with mother node M and 
daughters D$_1$ \dots D$_n$, generated by the corresponding rule
\verb|M ---> D1, ... DN.|, is formatted as \verb|M(D|$_1$, \dots,
\verb|D|$_n$). Daughters may again be local trees or, alternatively,
terminal symbols "T$_i$", the latter being simply formatted as T$_i$. 
Curly braces, denoting underparsing, may surround local trees.

Now, in general a constraint first reads in each candidate line and stores an
intermediate copy of the line in a buffer. It then looks
for occurrences of patterns that indicate constraint violation. The
offending pattern is substituted by asterisks, either directly or
through a cascade of intermediate substitutions. The latter option
often turns out to be a helpful {\em divide-and-conquer}-type strategy to
attack the problem at hand. Then all other material is 
deleted. The result constitutes a new column field in the violation
vector column under construction, with the position of the field
textually corresponding to the constraint's rank order, as is familiar from the
usual tableau layout. Appending the partial violation vector just
constructed to the previously saved copy of the candidate 
restores a complete candidate-plus-partial-annotation line, thereby
giving the next constraint in the rank order a chance as well. This
procedure is repeated for every candidate line from the GEN output.

The central question now is how to specify substitutions and which
mechanism to use for actually carrying out these pattern replacements.
A well-known efficient device for pattern substitution in linear time is a {\em
finite-state transducer} (FST). FSTs have well-understood mathematical
properties. In particular, they are closed under composition, which means
that for an arbitrary cascade of FSTs one can always find another
single FST that is completely equivalent in its behaviour.
The patterns that can be mapped via FSTs to other patterns
must be specifiable by a {\em regular expression}. In
OT SIMPLE Constraints-as-FSTs are specified in the format accepted by
`sed', a standard software tool that comes with the system software of
at least all UNIX OSes.
`sed' is a \underline{s}tream
\underline{ed}itor which can {\em inter alia} perform pattern
substitutions in the line-by-line manner outlined above.  In OT
SIMPLE, then, a constraint is made of a sequence of `sed' actions and
stored in a `sed' script file. 

The following is one the most simple constraints, \C{$\ast$Struc}
`Avoid structure' (cf. \citeNP[ch.3,fn.13]{prince:smolensky:93}). In the
formalization we adopt here it emits a violation 
mark for every immediate domination relationship in the tree. 
It follows that \C{$\ast$Struc} is an example of a binary constraint.
Luckily, in the 
flattened tree format described above there is a single-character
symbol indicating immediate domination in a local tree, the opening
round bracket.  This being said, let us look at the actual, unedited
formulation of the constraint below (note that empty and
hash sign-initial lines, the latter containing comments, are ignored by `sed'):
\begin{examples}
\item \label{NoStruc}
{\sc $\ast$Struc} \\
\footnotesize
\begin{verbatim}
# COMMON CONSTRAINT PROLOGUE
# save candidate to buffer

h

# delete violation vector (everything following the dot) 
# to get pure candidate

s/\..*//g

# CONSTRAINT-SPECIFIC MAIN PART
# delete everything except structural dominance symbol (

s/[^(]//g

# substitute ( by *

s/(/\*/g

# COMMON CONSTRAINT EPILOGUE
# remove all non-violation star material

s/[^\*]//g

# append violation stars after candidate (creates newline)

x;G

# convert superfluous newline into constraint separator character '

s/\n/\'/
\end{verbatim}
\end{examples}
Apart from the commands to save a copy of the current line (h),
exchange the current line buffer and the save buffer (x) and
append the saved copy after the current line (G), all other actions are
substitution specifications. Substitution commands will form the core
of all constraint specifications in this paper. Their concrete syntax is
\verb|s/Pattern/Replacement/g|. Note that the trailing \verb|g|
indicates that substitutions are to be made for every occurrence of
Pattern in the line, not just the first (for the latter option, omit
the \verb|g|). Space precludes a full explanation
of the syntax of admissible Pattern and Replacement expressions, for
which see e.g. \citeN[p.319-24]{gulbins:88}, but in order to understand the constraint
script in (\ref{NoStruc}) and others to follow, let us note a few
salient points here. 

While simple strings are typed in as such, the empty string denoting
deletion of the pattern is marked by a missing Replacement
specification, i.e. the second and third slashes are adjacent. If 
any of a whole set of characters should match
against a single character in the line, the set may be enumerated
between square brackets \verb|[]|. An initial circumflex \verb|^| acts as a
complement operator, i.e. with it a character is matched that is {\em
not} in the set of characters to follow. Some symbols including the
asterisk, the dollar sign and -- in 
some contexts -- also the round and square brackets have predefined meaning
for `sed'. If these must be matched literally, a preceding backslash
is provided as an escape character in order to turn off the
predefined meaning. One important predefined operator is the Kleene
star operator, written as an asterisk symbol. It is used for
specifying an unbounded number of concatenations (including zero
concatenation) of the character expression preceding it. The
predefined dollar symbol is also sometimes helpful, as it only matches
the end of the line. Finally, in deviance from the escape convention,
\verb|\n| matches the newline character. 

Since making the initial candidate copy and deleting old violation
stars will always form the prologue of any
constraint script, and deleting non-asterisk material, restoring the
copy and substituting the superfluous newline character by a single quote
always forms the epilogue, we shall omit both of these parts in the
text from now on. Their purpose is clear: provide a fresh start for
each constraint in the beginning and guarantee a well-defined
evaluation result at the end. In particular, an unviolated constraint
issues a single single quote character, whereas a constraint that is violated
N times emits a single quote followed by a contiguous string of N $>$
0 asterisks. 

In (\ref{Fill}) and (\ref{ParseSeg}) the faithfulness constraints {\sc
Fill} and {\sc Parse$_{Seg}$} are displayed, which penalize
epenthetic structure and segmental underparsing, respectively.
Unsurprisingly, they can be implemented as simple variants of {\sc
$\ast$Struc}, as the opening square and curly brackets similarly indicate the
presence of empty elements and underparsed input.%
\footnote{This constraint family relationship was already hinted at in
\citeN[fn.16]{mccarthy:prince:93}.}
\begin{examples}
\item \label{Fill}
{\sc Fill} \\
\begin{verbatim}
# delete everything except epenthesis symbol [

s/[^\[]//g

# substitute [ by *

s/\[/\*/g
\end{verbatim}
\end{examples}

\begin{examples}
\item \label{ParseSeg}
{\sc Parse$_{Seg}$}\\
\begin{verbatim}
# delete everything except underparse symbol {

s/[^{]//g

# substitute { by *

s/{/\*/g
\end{verbatim}
\end{examples}
A final remark: since constraints can become much more complex than
this, it is not only good style but almost vital for later constraint
maintenance to include a comment before each `sed' action.
\section{OT at work} \label{GenInAction}
It is time to put the pieces together and run through a very simple example.
Suppose we have prepared a correct GEN grammar file \verb|hessian.gen|
containing the material in (\ref{SampleGenGrammar}), and also stored the
above three constraints into script files \verb|NO-STRUC, FILL, PARSE-SEG|.
Then we are ready to start experimenting with OT SIMPLE.
In the following, the OS command interpreter prompt is displayed as a
dollar sign followed by a number, user keyboard input is in
\textsf{sans serif} typeface, and OT SIMPLE's response is in
\verb|typewriter| font. Witness the following interaction to just
generate the unevaluated candidate set from a simple input
\textsf{[t,a]} with a maximum of 2 epenthetic positions. The command
\textsf{GEN} is provided by OT SIMPLE, and its arguments must be
enclosed in double quotes, as shown.
\begin{examples}
\item \label{SimpleGeneration}
{\sc GEN at work}\\[2em]
\scriptsize
\texttt{\$1 }\textsf{GEN ''hessian, [t,a], 2''}
\begin{verbatim}
word(ft(syl(Rt(SPREAD_GLOTTIS,CORONAL),m(Rt(SONORANT,DORSAL))))).
word(ft(syl(m(Rt(SPREAD_GLOTTIS,CORONAL))),syl(m(Rt(SONORANT,DORSAL))))).
word(ft(syl(m(Rt(SPREAD_GLOTTIS,CORONAL)))),ft(syl(m(Rt(SONORANT,DORSAL))))).
word(ft(syl(Rt(SPREAD_GLOTTIS,CORONAL),m({Rt(SONORANT,DORSAL)})))).
...
word(ft(syl(Rt([]),m({Rt(SPREAD_GLOTTIS,CORONAL)})),syl(m(Rt([])))),\
    ft(syl(m({Rt(SONORANT,DORSAL)})))).
word(ft(syl(Rt([]),m(Rt([]))),syl(m({Rt(SPREAD_GLOTTIS,CORONAL)}))),\
    ft(syl(m({Rt(SONORANT,DORSAL)})))).
$2
\end{verbatim}
\end{examples}
This small example actually generates 432 candidates! Shown are the
first four candidates and the last two, in generation order. After the
first candidate displaying some reasonable structural assignment, i.e.
a single foot dominating a single syllable with /t/ in onset and /a/
dominated by a mora, things start getting weirder, with two syllables
and two feet being the next two possibilities. Then comes the first
candidate again, but with underparsed /a/ as indicated by curly
brackets around the root node subtree \verb|{Rt(SONORANT,DORSAL)}|.%
\footnote{The particulars of the (segmental) feature theory assumed
here are immaterial to the present discussion. Underparsing on the
level of root nodes stems from specifying root node rules
\texttt{SomeLabel \# `Rt' ---> \dots} as bearing the input labels in the
GEN grammar. With a different GEN grammar, the level at which
underparsing occurs can of course be changed at will.} 
The final two candidates all exhaust the maximum limit of two
epenthetic elements \verb|[]|, incorporated into different positions
of bizarre two-feet-three-syllable structures with both input segments
being underparsed. 

Let us move on to some constraint action.
Here is the result of plugging in CON with two constraints,
{\sc Parse}$_{seg}$ and $\ast${\sc Struc}. Note the use of the
predefined command \textsf{CON}, which takes a list of constraint
files in ranking order as arguments. As usual, individual OT
components which stand in a feeding relationship are connected via
pipes $|$. Two auxiliary commands provide some additional convenience.
\textsf{COUNT} augments the candidate display with line counts and
reformats the candidate line to display the violation vector below
each candidate for a more compact presentation.
\textsf{SHOW\_PAGEWISE} provides a  service for viewing the candidate
stream one screen page at a time (pressing the space bar displays the
next page). 
\begin{examples}
\item \label{SimpleConstraintAction}
{\sc GEN plus CON in action}\\[2em]
\scriptsize
\texttt{\$3 }\textsf{GEN ''hessian, [t,a], 2'' $|$ CON PARSE-SEG NO-STRUC
$|$ COUNT $|$ SHOW\_PAGEWISE}\\
\begin{verbatim}
1
word(ft(syl(Rt(SPREAD_GLOTTIS,CORONAL),m(Rt(SONORANT,DORSAL)))))
''******
2
word(ft(syl(m(Rt(SPREAD_GLOTTIS,CORONAL))),syl(m(Rt(SONORANT,DORSAL)))))
''********
3
word(ft(syl(m(Rt(SPREAD_GLOTTIS,CORONAL)))),ft(syl(m(Rt(SONORANT,DORSAL)))))
''*********
4
word(ft(syl(Rt(SPREAD_GLOTTIS,CORONAL),m({Rt(SONORANT,DORSAL)}))))
'*'******
...
\end{verbatim}
\end{examples}
Note that candidates 1-3 only display increasing amounts of
$\ast${\sc Struc} violations, whereas candidate 4 is the first to incur a
single {\sc Parse}$_{seg}$ violation due to /a/ underparsing. 

Finally, sticking in two more predefined commands, \textsf{EVAL},
for candidate sorting in increasing violation order, and \textsf{TREE},
a graphical tree display command operating on the topmost line,
we have assembled all that is necessary to display the optimal candidate:
\newpage
\begin{examples}
\item \label{SimpleEvalAction}
{\sc Computing the optimal output for} /ta/ \\[2em] 
\scriptsize
\texttt{\$4 }\textsf{GEN ''hessian, [t,a], 2'' $|$ CON PARSE-SEG NO-STRUC
$|$ EVAL $|$ TREE}\\
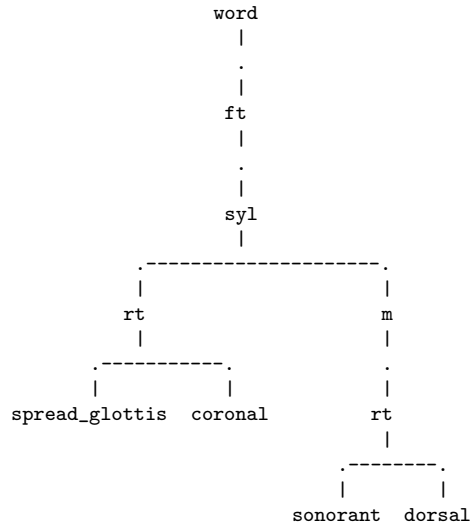
\begin{verbatim}
                                 word
                                   |
                                   .
                                   |
                                  ft
                                   |
                                   .
                                   |
                                  syl
                                   |
                          .---------------------.
                          |                     |
                         rt                     m
                          |                     |
                      .-----------.             .
                      |           |             |
               spread_glottis  coronal         rt
                                                |
                                            .--------.
                                            |        |
                                        sonorant  dorsal
\end{verbatim}
\end{examples}
The result is as expected: monosyllabic /ta/ as the winning candidate is
minimally structured and contains no underparsed segments.

In the next section we move on to something more serious, the task of
formalizing and implementing a real analysis that has been published. 
This case study serves to illustrate the practical utility of OT
SIMPLE, while also showing important additional constraint 
coding techniques.
\section{Formalizing Golston \& Wiese (1996)}
\citeN{golston:wiese:95} investigate what looks like a case of
subtractive plural morphology in Hessian German, illustrated in
(\ref{SubtractivePl}).
\begin{examples}
\item \label{SubtractivePl}
\begin{tabbing}
\underline{Singular} \= \underline{Plural\phantom{g}} \= \\
h\openo n\undercirc d 	\> h\openo n \> `dog' \\
d\openo\undercirc g 	\> d\openo\length \> `day' \\
b\niepsilon rg 		\> b\niepsilon r \> `mountain'
\end{tabbing}
\end{examples}
The authors develop an optimality-theoretic analysis on the basis of a broad range
of data including the items we have just seen, arguing that the
dialects of Hessian versus Standard German differ only in terms of
constraint ranking. Furthermore, they are able to dispense with the
assumption of a morphologically distinctive process of truncation for
the phenomenon at hand. In the following, let us provide a very brief
summary of the essentials of their analysis. The reader is urged to
consult the original paper for additional detail.
\subsection{The analysis of Hessian subtractive plural}
Golston \& Wiese note that, over and above the usual
\ling{-er,-n,-e,} -$\not {\! 0}$ plurals also found in Standard
German, Hessian has subtractive plurals. They summarize the conditions
under which subtraction occurs as follows: subtraction if the stem
ends in /ln, nd, \eng g, \scr g, Vg/. Hence, after omitting the last
segment in these sequences, such plurals end in a sonorant. This
observation makes some sense cross-dialectally, since in Standard
German plural nouns invariably end in an unstressed sonorant-final
(/\schwa,l,\invscr,n/) syllable.   
Importantly, in Hessian \mbox{-$\not {\! 0}$}~plural is in complementary
distribution with subtractive plural, the former occurring if the stem
does {\em not} end in /ln, nd, \eng g, \scr g, Vg/. 
The authors uncover the
generalization that subtractive plurals can only be formed if no
distinctive feature is lost in the process. Since they assume that
adjacent identical features within a morpheme are shared, stems
such as \ling{h\openo nd, d\openo g} are analyzed with
shared final coronal and dorsal features respectively. Last segment
subtraction delinks only one of the linked segments, hence preserving
the feature as a whole. Unlike these cases, stems such as \ling{flek}
`patch', \ling{h\niepsilon ls} `(no gloss)', \ling{\esh v\niepsilon nts} `(no gloss)'
have unshared final laryngeal and manner features (spread\_glottis and
continuant) under the featural classification scheme of Golston \&
Wiese (see (\ref{GenGrammarSegments}) below for a replication of this
scheme in OT SIMPLE's format).

Their constraint-based account of subtractive pluralization then boils
down to the following: Hessian plurals end in a sonorant, to be
captured by a constraint \textsc{Son]pl}.%
\footnote{See the original paper for an argument why \textsc{Son]pl} is
not 
a language-particular constraint.}
\textsc{Son]pl} must hold except when compliance with it would invoke
epenthesis or underparsing of distinctive features. This motivates additional
constraints \C{Fill} and \C{Parse-Feature} ranked above \C{Son]pl}.
The fact that segments are sometimes deleted in order to respect
\C{Son]pl} means that another constraint \C{Parse-Seg} must be ranked
below it. Segment-level underparsing unsurprisingly is the way to
achieve subtractive plurals, so familiar \C{Parse-Seg} ensures that
underparsing will be minimal.
\subsection{Implementing the analysis}
\subsubsection{Extending GEN}
A necessary ingredient of any implemented OT analysis is to get GEN to produce the
correct and all-inclusive range of structures for a given input. To
that end, let us extend the GEN structural grammar given earlier under
(\ref{SampleGenGrammar}). While the authors did not commit themselves to
a specific theory of prosodic structuring, taking over the particular
treatment in terms of feet, moras and root nodes from (\ref{SampleGenGrammar}) is
wellfounded. The simple reason why we can adopt this part of the GEN
grammar in unmodified form without fear of distorting their analysis is 
that no constraint used by Golston \& Wiese is sensitive to prosodic
structure.   
In lines 16 and 17 of (\ref{SampleGenGrammar}), then, we have already 
seen examples of the general format in which input-labelled rules for segmental
definitions can be specified. Extending the segmental definitions
beyond the two definitions for \verb|t,a|, we now
simply copy the distinctive feature chart of Golston \& Wiese into the
same rule format. This yields the addendum to \verb|hessian.gen|
depicted below:
\begin{examples}
\item \label{GenGrammarSegments}
\footnotesize
\begin{verbatim}
'O' # 'Rt' ---> "SONORANT", "DORSAL".                   %16
i   # 'Rt' ---> "SONORANT", "DORSAL".                   %17
'E' # 'Rt' ---> "SONORANT", "DORSAL".                   %18
u   # 'Rt' ---> "SONORANT", "DORSAL".                   %19
o   # 'Rt' ---> "SONORANT", "DORSAL".                   %20
                                                        %21
b   # 'Rt' ---> "LABIAL".                               %22
p   # 'Rt' ---> "SPREAD_GLOTTIS", "LABIAL".             %23
                                                        %24
d   # 'Rt' ---> "CORONAL".                              %25
t   # 'Rt' ---> "SPREAD_GLOTTIS", "CORONAL".            %26
                                                        %27
g   # 'Rt' ---> "DORSAL".                               %28
k   # 'Rt' ---> "SPREAD_GLOTTIS", "DORSAL".             %29
                                                        %30
s   # 'Rt' ---> "CONT", "SPREAD_GLOTTIS", "CORONAL".    %31
'S' # 'Rt' ---> "CONT", "SPREAD_GLOTTIS", "CORONAL".    %32
                                                        %33
n   # 'Rt' ---> "SONORANT", "NASAL", "CORONAL".         %34
m   # 'Rt' ---> "SONORANT", "NASAL", "LABIAL".          %35
'N' # 'Rt' ---> "SONORANT", "NASAL", "DORSAL".          %36
                                                        %37
l   # 'Rt' ---> "SONORANT", "CORONAL".                  %38
                                                        %39
r   # 'Rt' ---> "SONORANT", "DORSAL".                   %40
                                                        %41
h   # 'Rt' ---> "SPREAD_GLOTTIS".                       %42
\end{verbatim}
\end{examples}
Note, first of all, that due to restrictions in the machine
representation of IPA characters, IPA /\openo,\niepsilon,\esh,\eng/
appear as \verb|'O','E','S','N'|. Furthermore, definitions for vowels
lack distinctive place, as this was not given in Golston \& Wiese's
analysis, presumably because it is immaterial to plural formation
itself. In the same vein, definitions for \verb|'S',m,'N',r,h| were merely
inferred by analogy. 
With epenthesis
and underparsing already provided by earlier parts of the GEN grammar
and the underlying GEN mechanism, respectively, specification of GEN
is essentially complete. Producing a candidate set for, say,
\ling{h\openo nd} `dog', then will be as simple as \textsf{GEN
"hessian, [h,'O',n,d], 1"}.  
\subsubsection{The constraint \C{Son]pl}}
As a matter of fact, three of the five constraints needed to model
Golston \& Wiese's analysis are already in place:
\C{Parse$_{Seg}$} and \C{Fill} were defined earlier. The same goes for
\C{$\ast$Struc}, which did not appear in the original analysis, but
is entirely compatible with it and surely a necessary move to minimize
unmotivated prosodic structure. 

The remaining two constraints are \C{Parse-Feat} and \C{Son]pl}. They
are more difficult to implement than the other ones we have seen.
However, it turns out that they are also
quite instructive examples of important constraint specification
strategies that will be of general use in a regular transduction-based
constraint language.

Below comes \C{Son]pl}. We continue with our practice of leaving out
both the preparatory prologue and the cleaning-up actions of the
epilogue, both of which were previously identified as common to all
constraints. Also, since the extra morphological 
restriction to plural is obviously redundant in the context of our focussed
task of deriving only plurals, we refrain from introducing
additional complexity in GEN grammar, input specification and this
constraint, therefore omitting all actual reference to plural in the
implementation (as in Golston \& Wiese's paper). The reader is
encouraged to implement plural conditioning as an exercise. 
\enlargethispage{2\baselineskip}
\begin{examples}
\item \label{SonPl}
\C{Son]pl}\\[-\baselineskip]
\footnotesize
\begin{verbatim}
# delete uninteresting symbols

s/word//g;s/ft//g;s/syl//g;s/m//g;s/[\(\)]//g;s/[]\[]//g
s/CORONAL//g;s/NASAL//g;s/CONT//g;s/SPREAD_GLOTTIS//g
s/LABIAL//g;s/DORSAL//g;s/,//g

# start classifying four parsing states ...
#       substitute underparsed sonorant by 1

s/{RtSONORANT}/1/g

#       substitute parsed sonorant by 2

s/RtSONORANT/2/g

#       substitute underparsed non-sonorant by 3

s/{Rt}/3/g

#       substitute parsed non-sonorant by 4

s/Rt/4/g

# remove all underparsed segments, since interest here 
# is on PARSED sonorants

s/[13]//g

# insert speculative violation mark for word end $. Reason: we want
# totally underparsed words to violate the constraint as well!

s/$/\*/g

# revoke speculative violation mark * at right word 
# edge if preceded by parsed sonorant 2

s/2\*//g
\end{verbatim}
\end{examples}
As is typical for many constraints, \C{Son]pl} is only interested in
certain narrowly-defined configurations involving the feature
\verb|SONORANT|. Therefore, an initial step in (\ref{SonPl}) consists of
deleting all those uninteresting symbols that also populate the
candidate tree, in effect bringing the interesting ones closer
together. Note that in the absence of formal means to form a
complement set to a given set of interesting strings, an operation which is
unfortunately not available in the restricted inventory of regular
expression operators that `sed' implements, we must define the set of
deletable symbols by piecewise enumeration. This introduces a certain
amount of unnecessary interdependency between constraint formulation
and GEN specification, since quite general constraints that
could otherwise be reused with different GEN structural grammars here
must depend on a detailed knowledge of GEN.  

The next step illustrates another general technique: for the problem
at hand, identify a partitioning into meaningful equivalence classes
and substitute tokens of each of the classes by a canonical
representative, e.g. an otherwise unused natural number. While not
strictly needed from a formal point of view, 
substituting possibly complex tree configurations by short 
representatives certainly helps to keep constraint formulation manageable
for humans. Here our problem is how to identify parsed sonorants at the
right word edge. This induces a natural 2 x 2 partition with respect to
the binary dimensions {\em parsed} vs {\em underparsed} and {\em sonorant} vs
{\em non-sonorant}. Since underparsing itself was earlier defined to occur
at the root node level, we find expressions like \verb|s/{Rt}/3/g|,
which state that an underparsed non-sonorant (curly brackets
plus root node symbol plus absence of the string \verb|SONORANT|) is
to be substituted by \verb|3| wherever it occurs in the
entire candidate representation. 

Now, to understand the logic of the following lines it helps to 
consider the underparsing options in a bit more detail. If there is an
unbroken chain of final underparsed segments preceded by a parsed segment,
that segment would be final in terms of actual pronunciation,
since phonetic interpretation is deemed to disregard underparsed
structure. Its substantive content -- sonoranthood or not -- could
then be used to decide whether to output a violation mark or not.
However, a special case arises for totally underparsed 
input, the so-called `null parse' of \citeN{prince:smolensky:93}, where
there is no such left-over segment. Choosing the behaviour that seems
to be most conformant with the text analysis of Hessian, let us assume
that \C{Son]pl} is violated in this case. To correctly implement the
constraint, it is therefore best to proceed in an inverse fashion:
stipulate a violation and revoke it when the only positive
configuration, namely a parsed word-final sonorant, is met. 

In general, this `inverse' constraint coding
strategy is best used when enumerating all violating configurations
would be cumbersome, but describing the non-violating case is
comparably easy.
Towards this end we remove the underparsed segments, both to find the actual
word end and also because they do not contribute to our sought-after
positive configuration. Then the single stipulated violation mark is
inserted by transducing from the word end symbol that is invisibly contained in
any line of text according to `sed'. Since there will be no other
source of violation stars for this constraint, \C{Son]pl} is hereby
classified as a binary constraint, which is in accordance with the
published analysis although the authors do not overtly say so. 
If it was there in the candidate in the first place, we now have the
desired locally detectable context, namely a parsed-sonorant symbol
\verb|2| left-adjacent to the unique word-final violation symbol, which we 
now delete. All other configurations including the `null parse' do not
meet this context, hence the violation mark stays. 
\subsubsection{The constraint \C{Parse-Feat}}
The second constraint, \C{Parse-Feat}, shares some similarities with
the one we have just seen, but introduces additional interesting
complexities that are well-worth discussing. As it turns out, for
purposes of implementation it makes life easier to understand
\C{Parse-Feat} as a constraint {\em schema}, which is expanded into a
concrete constraint for each feature (cf. also \citeNP{wiltshire:94},
\citeNP{kiparsky:94}). What is therefore shown below is  
one concrete variant, \C{Parse-Coronal}, which may serve as a recipe
for all other incarnations of the schema. A second point worth noting
is that, at least for the analysis at hand, we can view feature-level
underparsing to follow from segment-level underparsing. This
asymmetric dependency trivially holds for isolated features of a given
underparsed segment, features that are absent
from adjacent segments. However, we will see that with appropriate
simulation of automatic sharing of adjacent features in the constraint
itself we can also reduce featural underparsing of shared instances to
segmental underparsing. Hence, the assumption of segment-level
underparsing, as specified in the current GEN structural grammar, need
not be revisited in what follows. Here then comes the formal implementation
of \C{Parse-Coronal}:
\newpage
\enlargethispage{2\baselineskip}
\begin{examples}
\item \label{ParseFeat}
\C{Parse-Coronal} \\[-\baselineskip]
\footnotesize
\begin{verbatim}
# delete uninteresting symbols

s/word//g;s/ft//g;s/syl//g;s/m//g;s/[\(\)]//g;s/[]\[]//g
s/SONORANT//g;s/NASAL//g;s/CONT//g;s/SPREAD_GLOTTIS//g
s/LABIAL//g;s/DORSAL//g;s/,//g

# start identifying four parsing states ...
#       substitute underparsed coronal by 1

s/{RtCORONAL}/1/g

#       substitute parsed coronal by 2

s/RtCORONAL/2/g

#       substitute underparsed non-coronal by 3

s/{Rt}/3/g

#       substitute parsed non-coronal by 4

s/Rt/4/g

# collapse adjacent 1...1 to 1 
# (reduce chain of underparsed coronals)

s/11*/1/g

# collapse adjacent 2...2 to 2 
# (effects automatic sharing of parsed coronals)

s/22*/2/g

# remove `protected' instances of underparsed coronals, i.e.
# adjacent to parsed coronals... parsed before underparsed order


s/21//g

#  ... underparsed before parsed order

s/12//g

# all remaining un'protected' underparsed coronals indicate constraint viol.

s/1/*/g
\end{verbatim}
\end{examples}
The upper portion of the constraint specification is by now familiar:
again, we see the removal of uninteresting symbols followed by a
four-way classification, this time relating to coronals. Only half of
the four classes are actually needed in the following, but we stick to the
general recipe for expository reasons.

The next stage is more interesting, as it shows the promised technique
for implementing cases of automatic, non-distinctive sharing in the
constraint itself, rather than having it already present in the
candidate. Recall that local sharing of features was a key ingredient
of Golston \& Wiese's analysis, forcing a need to reconcile this
demand with the tree structure limitation of candidates in OT SIMPLE.
Enriching the GEN implementation to cater for graph-structured output
invokes all sorts of questions relating to a suitable flat
representation of graphs that the constraints would be happy with,
questions we would like to put aside for the moment. Luckily, these
questions do not arise for automatic sharing under locality. The idea
is this: Whenever we have a contiguous chain of same-state features,
e.g. three parsed coronals in a row, we collapse the chain into a
single representative. This move then literally captures the essential
aspect of automatic sharing that we need here, namely that each shared
feature counts as only one instance on its tier. 

Now we are ready to identify distinctive feature losses. 
Note that under the intended behaviour of \C{Parse-Feature} demanded
by Golston \& Wiese's analysis we cannot
simply identify underparsed coronals with constraint violations. If
underparsed coronal segments are adjacent to a {\em parsed} coronal
segment, this corresponds to a scenario where at least one surviving
association line would still extend from the parsed segment on the
root node tier to the coronal tier, thereby `protecting' the coronal
feature. Therefore, such `protected' contexts do not count towards a
violation, and we need to delete them in order to properly isolate the
focal cases of unprotected underparsed coronals surrounded by non-coronals. 
These cases are then put into a one-to-one relationship with violation
marks, as expressed by the last line shown in (\ref{ParseFeat}). Again, the
epilogue not shown here deletes all non-asterisk material, thereby
including leftover instances of the classificatory numbers
\verb|2|\dots\verb|4|. 
\subsubsection{Running the analysis}
Having removed the last roadblock to an implemented version of
\citeN{golston:wiese:95}, let us put together the individual pieces to
show a full OT SIMPLE computation for the plural of \ling{h\openo nd} `dog':
\begin{examples}
\item \label{HondWinner}
{\sc Computing the plural} \\[2em]
\footnotesize
\textsf{\$5 GEN ''hessian, [h,'O',n,d], 1'' $|$ } \verb.\. \\
\textsf{CON PARSE-FEAT FILL SON]PL PARSE-SEG NO-STRUC $|$ EVAL $|$ TREE}
\end{examples}
(\ref{HondWinner}) states that \textsf{GEN} should generate output corresponding
to \ling{h\openo nd} with at most 1 epenthetic element. The output is fed
(\textsf{$|$}) into the constraint stage \textsf{CON}. \textsf{CON} takes the
constraint file names as argument, and delivers the concatenation of
the constraint scripts to `sed' for execution. The violation-annotated
output of this stage is then fed to \textsf{EVAL}, which sorts the
violation vectors. Finally, \textsf{TREE} picks the topmost-sorted
winning candidate and displays it in tree form. Here is this result,
the plural of \ling{h\openo nd} as actually output by OT SIMPLE: 
\begin{examples}
\item \label{HondWinnerTree}
{\sc The Winning Plural Candidate:} /h\openo n$<$d$>$/\\
\footnotesize 
\hspace*{-1cm}\begin{minipage}[t]{\linewidth}
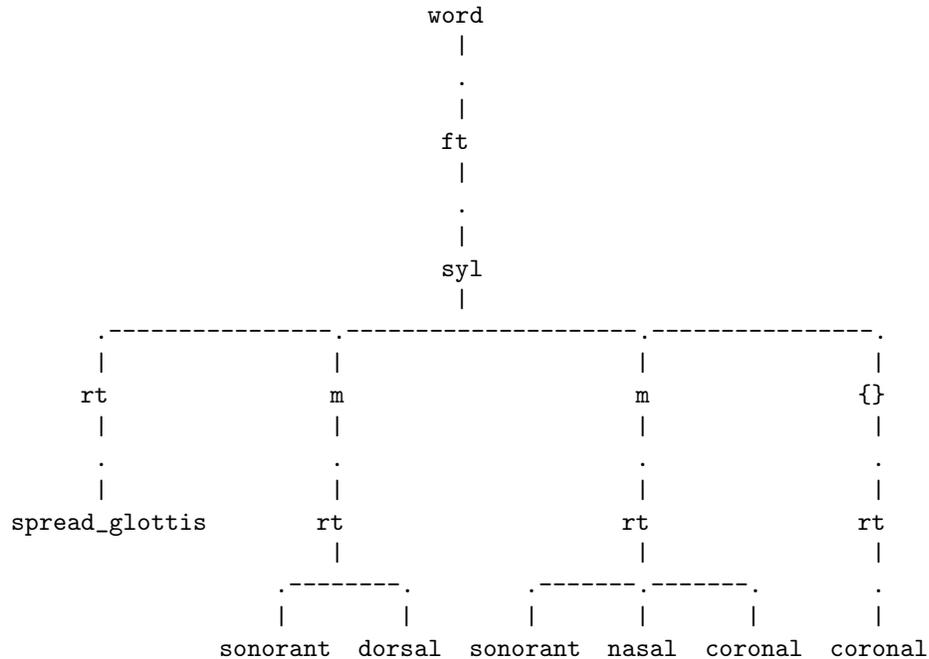
\begin{verbatim}
                                word
                                  |
                                  .
                                  |
                                 ft
                                  |
                                  .
                                  |
                                 syl
                                  |
        .----------------.---------------------.----------------.
        |                |                     |                |
       rt                m                     m               {}
        |                |                     |                |
        .                .                     .                .
        |                |                     |                |
  spread_glottis        rt                    rt               rt
                         |                     |                |
                     .--------.        .-------.-------.        .
                     |        |        |       |       |        |
                 sonorant  dorsal  sonorant  nasal  coronal  coronal
\end{verbatim}
\end{minipage}
\end{examples}
Note that the curly brackets which delimitated the underparsed
structure in the flat textual candidate representation reappear as a
designated tree node \verb|{}|.%
\footnote{This behaviour is due to Prolog, the programming language
used for GEN and TREE: its surface syntax allows a
convenient circumfix notation for curly brackets, in order to visibly
group together a possibly complex Prolog term. However, this is just
syntactic sugar for a unary functor symbol \texttt{\{\}}. Since the
tree output of general Prolog terms by \textsf{TREE} displays each
functor as a tree node and uses vertical arcs to indicate term
embedding, the picture follows.} 
The following is an excerpt from the full tableau containing the winner.
It shows the first three most harmonic candidates and the least
harmonic candidate, out of 2144 candidate lines and in the order
delivered by OT SIMPLE.%
\footnote{Actually, under the present GEN grammar there are four
candidates that tie for least harmony, since the epenthetic
element can be moraic or not and start a new foot or not. Both of these
dimensions are not subject to extra differentiating constraints in the
analysis that serves as our running example. When candidates have the
same harmony, their textual order as output by GEN is left unchanged
by EVAL/`sort', which behaviour classifies it as a {\em stable} sort.}   
\enlargethispage{-2\baselineskip}
\begin{examples}
\item \label{HondTableau}
{\sc Annotated Candidate Set for the Plural}\\
\scriptsize
\hspace*{-1cm}\begin{minipage}[t]{\linewidth}
\begin{verbatim}
1
word(ft(syl(Rt(SPREAD_GLOTTIS),m(Rt(SONORANT,DORSAL)),\
    m(Rt(SONORANT,NASAL,CORONAL)),{Rt(CORONAL)})))
''''*'*********
2
word(ft(syl(Rt(SPREAD_GLOTTIS),m(Rt(SONORANT,DORSAL))),\
    syl(Rt(SONORANT,NASAL,CORONAL),m({Rt(CORONAL)}))))
''''*'**********
3
word(ft(syl(Rt(SPREAD_GLOTTIS),m(Rt(SONORANT,DORSAL)))),\
    ft(syl(Rt(SONORANT,NASAL,CORONAL),m({Rt(CORONAL)}))))
''''*'***********
...
2144
word(ft(syl({Rt(SPREAD_GLOTTIS)},m({Rt(SONORANT,DORSAL)})),\
    syl(m({Rt(SONORANT,NASAL,CORONAL)}))),ft(syl(m(Rt([]))),syl(m({Rt(CORONAL)}))))
'**'*'*'****'****************
\end{verbatim}
\end{minipage}
\end{examples}
The topmost candidates incur just one \textsf{PARSE-SEG} violation, but
differ in their amount of structure as counted by \textsf{NO-STRUC} (9-11
violations). In contrast, the least harmonic candidate has 16
\textsf{NO-STRUC} violations, all four segments underparsed, violates
\textsf{SON]PL} and \textsf{FILL} once each, and contains two instances of
(coronal) \textsf{PARSE-FEAT} violation. Note that the last violation count
is explained by the epenthetic element that intervenes between coronal /n/ and
/d/, thus removing the adjacency required for automatic feature
sharing under the current formalization of \textsf{PARSE-FEAT}.

It is easy to see how, using OT SIMPLE, one could now go on to experiment with
different rankings by simply changing the order of \textsf{CON}
arguments, or single out specific constraints to observe their action
in isolation, or use a selected subset of salient candidates
previously stored in a file to speed up human inspection for purposes
of constraint debugging etc. 

One such case where quickly running an experiment with OT SIMPLE has
helped to uncover an important hidden assumption of
\citeN{golston:wiese:95} is the behaviour of Standard High German
\ling{h\scu nd} `dog' under pluralization, which the authors claim should be obtainable from
simply reranking \C{Fill} below \C{Son]pl} and \C{Parse-Seg}.
Surprisingly, however, we get the same output as for Upper Hessian instead of
the expected grammatical output \ling{h\scu nd\schwa}, with final schwa
epenthesis. A moment's reflection resolves the puzzle: the epenthetic
final element inserted by OT SIMPLE is devoid of featural content, contrary
to the tableau notation (\verb|[E]|) and wording (`epenthetic vowel') of
Golston \& Wiese. Therefore \C{Son]pl} cannot be satisfied. The small
point to be made here is that simply running the experiment in OT
SIMPLE was sufficient to point to the crucial nature of
the assumption of contentful epenthesis, something the authors did not
note {\em expressis verbis}. To remedy this defect in OT SIMPLE in a principled
manner, we will provide a suitable GEN facility to specify featural
content for epenthetic elements in the future. Meanwhile, a quick
fix would resort to enriching \C{Son]pl} by an additional initial
transduction step \verb|s/\[]/SONORANT/g|, thereby simulating
epenthetic sonoranthood on-the-fly. 

Summing up then, having a
software tool such as OT SIMPLE can be an effective means to 
uncover both minor and major defects of OT analyses that could all too
easily remain unspotted otherwise. The formal specification of GEN and
CON itself, a necessary first step before using OT SIMPLE to run analyses on
inputs, promises additional return value in terms of enhanced clarity
and precision in OT analyses.
\section{Implementation}
The following sections discuss some of the technical issues that were
important in implementing OT SIMPLE and seem particularly worth noting.
Together with the full source code listing in the appendix, these
remarks should help the knowledgeable reader to modify and extend OT
SIMPLE to suit his or her purposes.
They may be skipped by a reader who is only interested in the business
of formalizing and testing OT analyses on a computer. 
\subsection{Major ideas for implementing GEN in Prolog} \label{GenImplementation}
This section presupposes some basic knowledge in logic programming, in
particular Prolog.

GEN has been realized in a freely available variant of Prolog called
BinProlog (\citeNP{tarau:96}). As a platform for active research in logic
programming, BinProlog has nevertheless stabilized over the years,
evolving into a very useful programming language. There are four
properties that made it especially suited for GEN
implementation, as compared to other Prolog implementations: 

BinProlog 
\begin{itemize}
\item has a particularly efficient implementation of Definite Clause
Grammars (DCGs)
\item allows for multiple DCG input streams
\item contains linear logic constructs, especially linear assumption
\item provides for easy runtime system generation and C-ification
\end{itemize}
We will see the significance of each of these advantages later on.

The implementation of GEN  divides into two tasks. One is the compilation
of the external, human-readable notation for GEN grammars into an
internally usable format. The other consists of the generation process
itself, which is given an input and delivers the set of output
candidates for that input on the basis of a previously compiled GEN
grammar. Let us first describe the important aspects of the
compilation process in what follows. 

Iterating over the context-free rules in the GEN grammar, which is
consulted as an ordinary Prolog file, a
corresponding Prolog clause is asserted into the Prolog database for
each rule. However, rules fall into two classes depending on their
left-hand side (LHS). If a rule is labelled with an input marker
\texttt{Marker \# mother ---> daughters}, one must somehow ensure that
during the generation process the rule will be used only once for
every occurrence of \verb|Marker| in the triggering input. This is
markedly different from the desired behaviour for rules that have an
ordinary nonterminal as their LHS symbol, since these should in principle be
indefinitely reusable during a derivation. Therefore, although both
types of rules are converted into clausal form, the first type is
embedded under \texttt{\#/2} clauses with markers retained, while the second
type gives rise to predicate definitions where the LHS symbol directly
functions as the head functor.
This means that unlabelled rules-as-clauses are directly usable in a
consecutive generation process, while labelled clauses must be
activated depending on the input. The process of activation will be
explained below. 

A derivation starts from the distinguished startsymbol (if a
declaration to that effect was present) or from the default
startsymbol \verb|word| otherwise. Therefore the \verb|generate(Tree)|
definition that is compiled takes into account which of the two cases applies. 
The main result of a derivation is its derivation tree (also known as
a syntax or parse tree), built on the fly as a side effect of a successful
Prolog deduction starting 
top-down from the startsymbol goal. Therefore a distinguished argument
of the clause head corresponding to each rule is used to construct
that tree in Prolog term format, taking the LHS category as functor
while equating the term arity with the number of daughters on the right-hand
side of the rule (see Pereira \& Shieber 1987:74). 

The generation process caters for automatic
underparsing and bounded epenthesis. It must also make sure that inputs are
consumed only once during a derivation and that linear precedence
(LP) relations implicit in the ordered input are maintained. Finally,
all possible derivations consistent with the current GEN grammar must
be output for a given input, each derivation tree corresponding to a
candidate of the rich GEN candidate set.

One-time input consumption in this implementation corresponds to clauses that
are marked as being usable only once in a Prolog deduction, vanishing upon
backtracking. This is a nonstandard requirement that many current
Prolog versions cannot express directly. However, BinProlog provides a
predicate \verb|assumel/1|, for linear assumption, which has exactly
the desired effect. Using this predicate, one copy of the clausal form of a
labelled rule is linearly assumed for each corresponding input marker,
before actual generation starts.
To facilitate automatic underparsing, disjunctively a second copy is
assumed that differs only in its employment of a \verb|{}| wrapper
term. Ordering this disjunct second in the relevant predicate
\verb|assume_terminals_from_input| means that due to Prolog's SLD
proof strategy, candidates with more underparsing always appear later than
those with less underparsing. This is a desired result in the face of
typically limited amount of underparsing in actual winners --
candidates with minimal underparsing should therefore be found near the
beginning of the candidate stream in order to ease human inspection.

An important second requirement was to preserve the transitive LP relations
implicit in the input. Since GEN input is converted to linearly
assumed one-time clauses, thus existing on the level of grammar rules
during actual generation, no {\em technical} input has been necessary sofar.
Therefore, we are still free to use the DCG input stream for our own
purposes. The key idea is to reflect the LP relations%
\footnote{Taking our running example for illustration, in every
candidate for \texttt{[h,'O',n,d]}, \texttt{h} must be ordered left of
each of \texttt{'O',n,d}, \texttt{'O'} must precede \texttt{n,d} in the tree
yield and so on.}
by pairing GEN input positions with successive natural numbers, 
and setting the DCG input stream to the list of natural numbers thus obtained.
Extending our description of the compilation of linearly assumed
clauses corresponding to 
inputs, each such clause body is additionally specified to consume exactly one
position of the DCG input stream, which is required to unify with the
positional index that gave rise to it. E.g., in our \verb|[h,'O',n,d]|
example, the rule corresponding to \verb|n| can only be used in a
derivation if the next element that is not yet consumed is 3. Hence,
the top-down generation process is simultaneously one of parsing,
consuming the LP-encoding list. It is one of the virtues of
declarative programming languages like Prolog that such simultaneity of
top-down and bottom-up processing regimes requires no special
programming effort.

Given this much insight into the inner workings of GEN, it should be
quite easy to see now what bounded epenthesis corresponds to. We
simply treat epenthetic elements as another, but formally related,
type of input. This first of all means that we must linearly assume N
clauses with head \verb|empty([])| to model a case when exactly N
epenthetic elements are to be integrated into a tree derivation.
Perhaps more surprisingly, indexing epenthetic elements with natural
numbers exactly as in the previous case also makes sense. The reason is that
when repeatedly backtracking over a set of identical assumed clauses
Prolog would eventually try each possible matching order between rule
goals (in this case \verb|empty/1|) and stored clauses, leading to
uninformative spurious ambiguity. Imposing a total order on the
epenthetic pseudo-input to be respected by derivations effectively removes
such unwanted ambiguity.  Put another way, positional indexing of both
the GEN input proper and the epenthetic elements helps keep the GEN output
a true set, with multiple copies of the same element being disallowed. The only
difference of \verb|empty| clauses comes when consuming epenthetic
pseudo-input: we need to switch from the default DCG input stream to
a second DCG stream for epenthetic positions, consume one element that
must unify with the stored index position, then switch back to the
default stream. This ability to switch back and forth between multiple
DCG inputs is a distinct advantage of BinProlog's nonstandard
implementation of Definite Clause Grammars. We will see another
potential application of input stream switching in section
\ref{MultipleInputStreams}. 

Putting the pieces together is now really simple. A distinguished
predicate \verb|gen_workhorse| fills both DCG input streams with their
respective index lists, as derived from GEN input and the
user-specified number of epenthetic elements. Then it tries to prove
the goal \verb|generate(Tree)| whose body contains the start symbol of the
GEN grammar, using the standard depth-first left-to-right proof
strategy of Prolog. The proof is deemed successful if both input
streams have been fully consumed, instantiating \verb|Tree| as a
side-effect, which can then be output. In a failure-driven loop, all
different ways of proving a derivation from the start symbol are
found, with concomitant candidate output contributing to the richness
of GEN. Another iteration on top 
of this workhorse predicate enumerates the epenthetic possibilities,
starting from no epenthesis to the maximal number of epenthetic
elements that was specified by the user. 

An additional technical quirk is due to special BinProlog behaviour in
connection with assumed clauses. Since assumed clauses override all
compiled and asserted clauses with the same head, an unlabelled rule
with the same LHS as a labelled one would under normal assertion into
the Prolog database not be visible if a corresponding marker label
appeared in the GEN input. Therefore we need to lift these unlabelled
rules to the same level as labelled ones to prevent invisibility due
to overriding. Fortunately, a twin predicate of linear assumption
does exists in BinProlog, \verb|assumei/1| for intuitionistic assumption,
which in contrast to the linear logic construct can be used an
indefinite number of times in proofs, but also vanishes on
backtracking. We therefore modify the compilation of these unlabelled
rules accordingly by deferring actual clause assertion
in favour of recording a `promise' to activate clauses.
Activation then simply amounts to using the `promise' facts  at actual
generation time to intuitionistically assume each of the clauses so
promised, with the effect that same-LHS labelled and unlabelled rules can coexist as desired.

Finally, BinProlog provides a variety of ways to turn Prolog
programs into standalone applications, in particular by C-ification of both
the program and the necessary parts of BinProlog. This means that we
can turn the GEN implementation into a directly executable piece of
software which, when constructed carefully, can run on systems that
do not have BinProlog installed. See the BinProlog manual (\citeNP{tarau:96})
for details. 
\subsection{Issues in implementing constraint action}
The basic idea of implementing CON was very simple, using `sed'
scripts to specify finite-state-transducers that map selected portions of
candidate trees to violation marks. A substantial point not discussed
so far is the severity of the limitation imposed by the regularity
assumption for constraints, a limitation that is shared by all current
work that is explicit about formal constraint specification.

\citeN{ellison:95} has shown that the widely used family of Generalized Alignment
constraints (GA, \citeNP{mccarthy:prince:93a}) cannot be described with regular
language expressions, in contrast to the constraints we have seen so
far. While such qualitative difference in formal language class in an
innocuously-looking constraint family such as GA is certainly
surprising from the viewpoint of informal OT, the result obviously
needs further investigation in the present context. 
In fact, there is some hope that GA constraints
can be modelled using `sed', because some of its operations that were hitherto
unused in the main part of constraints go beyond regular power (a
case to remember is the {\em copying} of a previously stored candidate of a
priori unknown length, to be later inserted before the violation
vector in the constraint epilogue).  

Also it is natural to ask whether
the notion of  {\em correspondence} and its associated constraint
families (\citeNP{mccarthy:prince:94} et alia) can be incorporated into OT
SIMPLE. While this interesting problem must be left for further
research, a general idea that suggests itself is to compactify
correspondence pairs (or even N-tuples of correspondents) via some
sort of difference structures, in order to retain a sensible
definition of locality of corresponding elements for the constraints.  
\subsection{Issues in the implementation of EVAL}
We have seen that simple sorting suffices to model EVAL, here narrowly
understood as denoting the harmony maximization part proper.
However, there is room for improvement, since we did not take any
actions to {\em prune} the set of candidates after each evaluation of a
constraint, corresponding to the `!' graphical device used in OT
tableaux to indicate fatal violation. 
Recall that pruning exploits the familiar characteristic of EVAL that,
due to OT's strict domination regime, at any point in the constraint
hierarchy only those output candidates need to be retained that still
tie for optimality. 

We have in fact experimented with a shell script implementation of such a PRUNE
operator, constructed using only standard UNIX tools. The idea was to
sort the incoming candidate set on the basis of the partial violation
vector constructed so far, then examine the first line of the sorted
candidate stream to extract the currently optimal violation vector.
After converting the vector into regular expression format, we could then
filter the candidate stream with the UNIX tool `egrep', using the
regular expression just constructed, thereby literally deleting
non-optimal candidate lines. Due to the overhead incurred from shell script
interpretation and the usage of generic tools, however, the use of
PRUNE after each constraint evaluation turned out to be actually
slower, at least for small-scale examples. In order to be able to process huge
candidate sets, in the long run a more efficient implementation of
PRUNE is desirable. 

Finally, there are cases when the constraint hierarchy is not a total
ordering relation over CON. Since we can compose any partial order
from the union of suitable total ordering relations, it currently
seems best to manually form the set union of the winners corresponding
to each total order, as computed by OT SIMPLE. Of course, this step
would be easy to automatize if desired.
%
%
\section{Extensions}
For the immediate future we foresee two specific extensions that are
already known to be relatively easy to implement. We motivate both of
them in the following and sketch their implementation.
\subsection{Prespecification of structure}
Prespecification%
\footnote{Thanks to Chris Golston for suggesting this extension, prompted by a
demonstration where hundreds of output candidates flooded the computer screen.}
of structure that would be assigned anyway in some candidates
by GEN is an effective means to limit the size of the GEN set over
which CON and EVAL must operate, with obvious practical advantages for
analysis construction and debugging. It can also be seen as a facility
for implementing the {\em results} of Lexicon Optimization -- a procedure to
infer optimal inputs from surface forms
(\citeNP[ch.9]{prince:smolensky:93}) --, since \citeN{inkelas:94} has
pointed out an important connection between Lexicon Optimization and the possibility of
prespecifying predictable non-alternating structure underlyingly.

The implementation of prespecification capitalizes on the possibility
of specifying partial trees as partially instantiated
Prolog terms. Since \verb|generate(Tree)| returns a fully instantiated
tree after derivation, unifying the \verb|Tree| with the prespecified partial
tree before generation should be an effective forward-checking way of
enforcing the restrictions imposed by prespecification.
The question that remains then is what a suitable formal language should look
like that enables users to describe such incomplete trees
in an intuitive fashion. One good candidate would be a kind of modal
language, using prefix operators \verb|up,down,left,right| to wander around
the tree from some designated starting node, while mentioning restrictions on
the identity of the category nodes that are visited. A hypothetical
example expression in such a language might be 
\verb|right syl up ft left ft up up phrase| (`to my right, I see a syllable that is
dominated by a foot whose left-hand sister is another foot that itself
is dominated by a phrase located two nodes above').  
\subsection{Multiple input streams} \label{MultipleInputStreams}
We have already seen that internally the GENeration mechanism makes
use of two distinct DGC input streams that correspond to real GEN input (or, more
precisely, the list of input positions) and epenthetic pseudo-input.
This was possible because of the general provision of up to 255
distinct input streams in BinProlog, between which one can switch back
and forth at will in a DCG application.

Now, for cases like the celebrated Tagalog \ling{um} infixation analysis
of \citeN[ch.4]{prince:smolensky:93} and similar instances of mobile or
floating morphemes that can be analyzed as underspecified w.r.t.
linear precedence, this BinProlog feature again seems entirely
appropriate. Each such floating element would be assigned its
own ordered input stream, and the GEN mechanism would pick
nondeterministically from the totality of input streams arising from
the current GEN input, just as it nondeterministically selected
among parsed and underparsed variants of input items. While thereby
all possible intercalations of input streams would be output, the
order of each individual input stream could still be preserved by
positional indexing, as detailed above. With more articulate
categorial information in both the appropriate input and GEN grammar
specification -- at least including morphological tags to distinguish
e.g. stems from affixes -- constraints could then be straightforwardly
implemented to handle Tagalog \ling{um} and similar analyses.
\section{Conclusion}
We have described a particularly simple approach to the computer implementation of
classical OT. It is the first we know of that is publicly available,
yet offers generic mechanisms to specify constraints and GEN structural
grammars while simultaneously being geared towards practical usage.
Simplicity in implementation in particular stems from reusing standard
system software for CON and EVAL as well as from choosing BinProlog as a suitable
programming platform for GEN.
Due to the level of formal detail involved, working with OT SIMPLE may present initial
difficulties to some, specifically when it comes to the task of formalizing
constraints. Still, the overall behaviour and look-and-feel of this software tool
should be close enough to the original presentation of OT to make OT
SIMPLE fairly intuitive to use. It is therefore hoped that OT 
enthusiasts will indeed employ it to develop and test implemented and
formally sound OT analyses, both for their own merit and possibly also in the
classroom, invent new constraint coding techniques beyond those
described and suggest further improvements and extensions. 
\bibliographystyle{chicago}

\begin{thebibliography}{}

\bibitem[\protect\citeauthoryear{Andrews}{Andrews}{1995}]{andrews:95}
Andrews, A. (1995).
\newblock {OT} for {W}indows 1.1.
\newblock (ROA-91).
\newblock Software \& Documentation.

\bibitem[\protect\citeauthoryear{Bird}{Bird}{1991}]{bird:91}
Bird, S. (1991).
\newblock {F}eature {S}tructures and {I}ndices.
\newblock {\em Phonology\/}~(8), 137--144.

\bibitem[\protect\citeauthoryear{Ellison}{Ellison}{1994a}]{ellison:94a}
Ellison, T.~M. (1994a).
\newblock {C}onstraints, {E}xceptions and {R}epresentations.
\newblock In {\em Proceedings of ACL SIGPHON First Meeting}, pp.\  25--32.

\bibitem[\protect\citeauthoryear{Ellison}{Ellison}{1994b}]{ellison:94}
Ellison, T.~M. (1994b).
\newblock {P}honological {D}erivation in {O}ptimality {T}heory.
\newblock In {\em Proceedings of COLING'94}, Volume~II, pp.\  1007--1013.

\bibitem[\protect\citeauthoryear{Ellison}{Ellison}{1995}]{ellison:95}
Ellison, T.~M. (1995).
\newblock {OT}, {F}inite-{S}tate {R}epresentations and {P}rocedurality.
\newblock In {\em Proceedings of the Conference on Formal Grammar}, Barcelona.

\bibitem[\protect\citeauthoryear{Golston and Wiese}{Golston and
  Wiese}{1996}]{golston:wiese:95}
Golston, C. and R.~Wiese (1996).
\newblock {Z}ero morphology and constraint interaction: subtraction and
  epenthesis in {G}erman dialects.
\newblock In G.~Booij and J.~van Marle (Eds.), {\em Yearbook of Morphology
  1995}, pp.\  143--159. Kluwer Academic Publishers.
\newblock (ROA-100, posted 1995).

\bibitem[\protect\citeauthoryear{Gulbins}{Gulbins}{1988}]{gulbins:88}
Gulbins, J. (1988).
\newblock {\em {UNIX}. Eine {E}inf\"uhrung in {B}egriffe und {K}ommandos von
  {UNIX} -- {V}ersion 7, bis {S}ystem {V}.3}.
\newblock Springer.

\bibitem[\protect\citeauthoryear{Hammond}{Hammond}{1995}]{hammond:95}
Hammond, M. (1995).
\newblock {S}yllable parsing in {E}nglish and {F}rench.
\newblock Ms., University of Arizona, Tucson.(ROA-58).

\bibitem[\protect\citeauthoryear{Hopcroft and Ullman}{Hopcroft and
  Ullman}{1979}]{hopcroft:ullman:79}
Hopcroft, J. and J.~Ullman (1979).
\newblock {\em {I}ntroduction to {A}utomata {T}heory, {L}anguages and
  {C}omputation}.
\newblock Reading, MA: Addison-Wesley.

\bibitem[\protect\citeauthoryear{Inkelas}{Inkelas}{1994}]{inkelas:94}
Inkelas, S. (1994).
\newblock {T}he consequences of optimization for underspecification.
\newblock Ms., University of California, Berkeley.(ROA-40).

\bibitem[\protect\citeauthoryear{Kiparsky}{Kiparsky}{1994}]{kiparsky:94}
Kiparsky, P. (1994).
\newblock {R}emarks on markedness.
\newblock Handout of talk presented at TREND-2, Jan. 22.

\bibitem[\protect\citeauthoryear{McCarthy and Prince}{McCarthy and
  Prince}{1993}]{mccarthy:prince:93}
McCarthy, J. and A.~Prince (1993).
\newblock {P}rosodic {M}orphology {I}: {C}onstraint {I}nteraction and
  {S}atisfaction.
\newblock Technical report RuCCS-TR-3, Rutgers University Center for Cognitive
  Science.
\newblock (to appear, MIT Press).

\bibitem[\protect\citeauthoryear{McCarthy and Prince}{McCarthy and
  Prince}{1994a}]{mccarthy:prince:93a}
McCarthy, J. and A.~Prince (1994a).
\newblock {G}eneralized {A}lignment.
\newblock In G.~Booij and J.~van Marle (Eds.), {\em Yearbook of Morphology
  1993}, pp.\  79--153. Dordrecht: Kluwer.
\newblock (Technical Report \# 7, Rutgers University Center for Cognitive
  Science, 1993.ROA-7).

\bibitem[\protect\citeauthoryear{McCarthy and Prince}{McCarthy and
  Prince}{1994b}]{mccarthy:prince:94}
McCarthy, J. and A.~Prince (1994b).
\newblock {T}he {E}mergence of the {U}nmarked: {O}ptimality in {P}rosodic
  {M}orphology.
\newblock In M.~Gonzalez (Ed.), {\em Proceedings of the North-East Linguistics
  Society}, Number~24, Amherst, MA, pp.\  333--379. Graduate Linguistic Student
  Association. (ROA-13).

\bibitem[\protect\citeauthoryear{Prince and Smolensky}{Prince and
  Smolensky}{1993}]{prince:smolensky:93}
Prince, A. and P.~Smolensky (1993).
\newblock {O}ptimality {T}heory. {C}onstraint {I}nteraction in {G}enerative
  {G}rammar.
\newblock Technical report RuCCS TR-2, Rutgers University Center for Cognitive
  Science.
\newblock (To appear, MIT Press).

\bibitem[\protect\citeauthoryear{Raymond and Hogan}{Raymond and
  Hogan}{1996}]{raymond:hogan:96}
Raymond, W. and A.~Hogan (1996).
\newblock {A} users guide to the {O}ptimality {I}nterpreter: A software tool
  for {O}ptimality {T}heoretic analysis.
\newblock Ms., University of Colorado at Boulder.(ROA-130).

\bibitem[\protect\citeauthoryear{Scobbie}{Scobbie}{1991}]{scobbie:91}
Scobbie, J.~M. (1991).
\newblock {T}owards {D}eclarative {P}honology.
\newblock In S.~Bird (Ed.), {\em Declarative Perspectives on Phonology},
  Volume~7 of {\em Edinburgh Working Papers in Cognitive Science}, pp.\  1--26.
  Centre for Cognitive Science, University of Edinburgh.

\bibitem[\protect\citeauthoryear{Scobbie, Coleman, and Bird}{Scobbie
  et~al.}{pear}]{scobbie.et.al:96}
Scobbie, J.~M., J.~S. Coleman, and S.~Bird ({to appear}).
\newblock {K}ey {A}spects of {D}eclarative {P}honology.
\newblock In {\em Proceedings of Current Trends in Phonology Meeting '95}.
  Royaumont, France.

\bibitem[\protect\citeauthoryear{Tarau}{Tarau}{1992}]{tarau:92}
Tarau, P. (1992).
\newblock {B}in{P}rolog: a {C}ontinuation {P}assing {S}tyle {P}rolog {E}ngine.
\newblock In M.~Bruynooghe and M.~Wirsing (Eds.), {\em Proceedings of
  Programming Language Implementation and Logic Programming}, Number 631 in
  Lecture Notes in Computer Science, pp.\  479--480. Springer.

\bibitem[\protect\citeauthoryear{Tarau}{Tarau}{1996}]{tarau:96}
Tarau, P. (1996).
\newblock Bin{P}rolog 5.21 {U}ser {G}uide.
\newblock Technical Report 96-1, D\'{e}partement d'Informatique, Universit\'{e}
  de Moncton.
\newblock Available from {\em http://clement.info.umoncton.ca/BinProlog}.

\bibitem[\protect\citeauthoryear{Tesar}{Tesar}{1994}]{tesar:94}
Tesar, B. (1994).
\newblock {P}arsing in {O}ptimality {T}heory: A dynamic programming approach.
\newblock Technical Report CU-CS-714-94, Department of Computer Science,
  University of Colorado, Boulder.

\bibitem[\protect\citeauthoryear{Tesar}{Tesar}{1995}]{tesar:95}
Tesar, B. (1995).
\newblock {C}omputing optimal forms in {O}ptimality {T}heory: Basic
  syllabification.
\newblock Ms., University of Colorado and Rutgers University.(ROA-52).

\bibitem[\protect\citeauthoryear{Tesar}{Tesar}{1996}]{tesar:96}
Tesar, B. (1996).
\newblock {C}omputing {O}ptimal {D}escriptions for {O}ptimality {T}heory
  {G}rammars with {C}ontext-{F}ree {P}osition {S}tructures.
\newblock In {\em Proceedings of the 34th Annual Meeting of the ACL}.

\bibitem[\protect\citeauthoryear{Wiltshire}{Wiltshire}{1994}]{wiltshire:94}
Wiltshire, C. (1994).
\newblock {T}he need for {P}arse-feature constraints. {LSA} paper.
\newblock Ms., Brown University.

\end{thebibliography}

\newpage
\appendix
\section{Appendix}
This section contains first of all the source code for GEN, written in
BinProlog and contained in the current OT SIMPLE distribution as file
\verb|gen.pl|. The following three shell scripts encapsulate technical
details of the actual GEN, CON and EVAL invocations, which provide
the user interface to OT SIMPLE that was described earlier.
\paragraph{\Large gen.pl}
\footnotesize
\begin{verbatim}
%%%%%%%%%%%%%%%%%%%%%%%%%%%%%%%%%%%%%%%%%%%%%%%%%%%%%%%%%%%%%%%%%%%
%%%                                                             %%%
%%%         File: gen.pl                                        %%%
%%%       Author: Markus Walther                                %%%
%%%      Purpose: Model GEN component of OT in BinProlog        %%%
%%%    Copyright: Seminar fuer Allgemeine Sprachwissenschaft    %%%
%%%               Universitaet Duesseldorf, 1996                %%%
%%%                                                             %%%
%%%%%%%%%%%%%%%%%%%%%%%%%%%%%%%%%%%%%%%%%%%%%%%%%%%%%%%%%%%%%%%%%%%

%%% LOAD: $ bp -q 6
%%%       ?- compile(gen).


%%% KNOWN DEFICIENCIES
/*
        - empty positions can not be underparsed.
        - no explicit sharing
        - Multiple identical labels don't work: textually later rules
        are not catered for during input processing; e.g. in

        l # ons ---> "l".
        l # coda ---> "l".

        the second marked rule is never used.

*/

%%% OPERATORS

:- op(1200,xfy,('#')).
:- op(1150,xfx,('--->')).
:- op(900,fy,(gen)).
:- op(900,fy, (genload)).
:- op(900,fy, (startsymbol)).


%%%
% A SIMPLE REALIZATION OF OT'S GEN
%%%

% --- interactive GEN

gen(Input) :-
  write('Enter max. number of epenthetic positions, followed by period: '),
  read(NumberOfEpenthetic),
  gen(Input, NumberOfEpenthetic).

% --- non-interactive GEN
% --- enumerates candidates with increasing number of epenthetic elements

gen(Input, MaxNumberOfEpentheticPositions) :-
        MaxNumberOfEpentheticPositions >= 0,
        nl,
        for(CurrentNumberOfEmptyElements, 0, MaxNumberOfEpentheticPositions),
        gen_with_empty_structure(CurrentNumberOfEmptyElements, Input),
        fail.
gen(_,_) :- nl.

% --- GEN with structural grammar preload

gen(GrammarFile, Input, MaxNumberOfEpentheticPositions) :-
        genload(GrammarFile),
        gen(Input, MaxNumberOfEpentheticPositions).

gen_with_empty_structure(Exactly_N_Empties, Input) :-  
        assume_n_empties(Exactly_N_Empties, Exactly_N_EmptyPosList),
        assume_terminals_from_input(Input, PosList, 1),
        assume_static_terminals_from_grammar,
        gen_workhorse(PosList, Exactly_N_EmptyPosList).

assume_static_terminals_from_grammar :-
        retract(to_be_assumed(Clause)),         % fetch Clause from database
        assumei(Clause),                        % intuitionistic assume
        assume_static_terminals_from_grammar,   % recurse
        assertz(to_be_assumed(Clause)),         % re-store Clause in database
        !.
assume_static_terminals_from_grammar.

assume_terminals_from_input([],[], _) :- !.
assume_terminals_from_input([Marker|Markers], [I|PosList], I) :-
        ( ( Marker # (LHS :- RHS) ) -> true
        ; write('%%% ERROR: Input element '), write(Marker), 
          writeln(' triggers (#) no rule in the grammar!'),
          abort
        ),
        ( assumel((LHS :- RHS, dcg_connect(I)))
        ; functor(LHS, F, 1), functor(UnderparsedLHS, F, 1),
          arg(1, LHS, LHSArg), arg(1, UnderparsedLHS, {LHSArg}),
          assumel((UnderparsedLHS :- RHS, dcg_connect(I))) %%% UNDERPARSING
        ),
        J is I + 1,
        assume_terminals_from_input(Markers, PosList, J).

assume_n_empties(0,[]) :- !, assumel((empty(_) :- fail)). %% suppress 
                                                          %% undefined empty/1
                                                          %% warnings
assume_n_empties(N,[N|More]) :-                           
        M is N-1,                                         
        assumel((empty([]) :- dcg_tell(2), % switch to epenth. elem. stream
                              dcg_connect(N), % consume one elem.
                              dcg_tell(1))), % switch back to input stream
        assume_n_empties(M,More).

gen_workhorse(PosList, EmptyPosList) :- 
        dcg_tell(2),                    % switch to epenth. elem. stream
        dcg_def(EmptyPosList),          % define list of numbered epenth.elems
        dcg_tell(1),                    % switch to real input stream
        dcg_def(PosList),               % define input positions to be parsed
        generate(Tree),                 % run DCG, thereby instantiating Tree 
        dcg_tell(1),                    % check that input 
        dcg_val([]),                    % has been consumed, i.e. parse success
        dcg_tell(2),                    % check that all epenthetic elems
        dcg_val([]),                    % have been parsed/integrated as well
        gen_output(Tree),               % output Tree
        fail.                           % backtrack

gen_output(Tree) :- 
        write(Tree),                    % tree in Prolog term repres,
        put(46),                        % followed by single dot;
        nl.                             % each tree on separate line

initialize :-
        abolish_dynamic_predicates.

abolish_dynamic_predicates :-
        current_predicate(Pred,PredTerm),
        predicate_property(PredTerm, (dynamic)),
        functor(PredTerm, _Functor, Arity),
        abolish(Pred, Arity),
        fail.
abolish_dynamic_predicates.

%%
% PREDICATES BELONGING TO GEN GRAMMAR FILE COMPILATION
% A Gen grammar is compiled into a DCG in efficient hidden accumulator format
% provided by BinProlog. The DCG constructs a syntax tree as a side effect 
% while parsing input. This syntax tree is the main result for
% purposes of constraint evaluation, which itself is done outside of Prolog.
%%

% --- expand surface notation to internal

term_expansion( (Head0 ---> Body0), (Head :- Body)) :-
        !,
        compute_rhs(Body0, Body, BodyVars, BodyLength),
        functor(Head1, Head0, BodyLength),
        compute_lhs(BodyVars, Head1, 1),
        functor(Head, Head0, 1),
        arg(1, Head, Head1),
        !.

% --- compute right hand side (rhs) of rule

compute_rhs((First0,Rest0), (First,Rest), [FirstArg|RestArgs], Len) :-
        !,
        process_term(First0, First, FirstArg),
        compute_rhs(Rest0, Rest, RestArgs, RestLen),
        Len is RestLen + 1.

compute_rhs(Single0, Single, [SingleArg], 1) :-
        process_term(Single0, Single, SingleArg).

% --- process individual item in rhs of a rule

process_term(Single0, Single, SingleArg) :-
        ( atomic(Single0) -> 
                functor(Single, Single0, 1),
                arg(1, Single, SingleArg)
        ; name(SingleArg, Single0), Single = true
        ; writeln('%%% ERROR: right hand side categories in rule can only be'),
          writeln('%%% atoms, numbers or strings! Found "'),
          write(Single0), write('" instead.'), 
          Single = grammar_error(SingleArg), 
          SingleArg=here
        ).

% --- compute left hand side (lhs) of rule

compute_lhs([], _, _) :- !.
compute_lhs([FirstVar|RestVars], Head, I) :-
        arg(I, Head, FirstVar),
        J is I+1,
        compute_lhs(RestVars, Head, J).

translate_gen_grammar :- 
        ( (H ---> B), NonVarMarker=true % unmarked rule
        ; '#'(Marker, (H ---> B)),      % rule labelled with Marker
          ( nonvar(Marker) -> NonVarMarker=true
          ; NonVarMarker=false
          )
        ),
        ( ( call(NonVarMarker),                  % Marker was wellformed
            term_expansion(( H ---> B ), Clause) % translate rule to internal
          )  ->                         
                put(46)                          % progress report
        ; writeln('%%% WARNING: The following rule could not be translated '),
          writeln('%%% '), write((H ---> B)),  
          Clause = '$no_legal_rule'
        ),
        ( atomic(Marker) ->             % a marked rule
                assertz((Marker # (Clause))) % assert marker+rule translation
        ; '#'(_Marker, (H ---> _)) ->   % an unmarked rule for which a marked
                                        % one with same lhs exists
                assertz(to_be_assumed(Clause)) % mark rule translation as 
                                               % special since assumed preds 
                                               % override dynamic ones
        ; assertz((Clause))             % default: unmarked rule, enter transl.
        ),
        fail.                           % backtrack

translate_gen_grammar :-        
        ( startsymbol(Start) ->         % explicit startsymbol decl. exist
                functor(Startsymbol, Start, 1),         % construct start 
                arg(1, Startsymbol, Tree),              % category;
                assert((generate(Tree) :- Startsymbol)) % enter start pred.

        ; assert((generate(Tree) :- word(Tree))),       % enter default start
          writeln('%%% WARNING: No explicit startsymbol declaration found.'), 
          writeln('%%%          Assuming "startsymbol word." as default.') 
        ).

genload(GenGrammarfile) :-
  initialize,                   
  (             % construct full file name  from abbreviated GenGrammarfile
    name(GenGrammarfile, FileString0),
                %            .   g   e    n   filename extension
    det_append(FileString0, [46,103,101,110], FileString),
    name(File, FileString),
    consult(File) ->
                        write('%%% Translating Gen grammar file '),
                        ( translate_gen_grammar ->
                                nl, writeln('%%% Gen translated.')
                        ; nl, writeln('%%% Translation error.')
                        )
  ; writeln('%%% Gen grammar file name (*.gen) incorrect or file not found')
  ), 
  nl.
        
help :- 

writeln('%%%                   OT SIMPLE / GEN version 1.0'),
writeln('%%%                   ---------------------------'),
writeln('%%% (c) 1996 Markus Walther (walther@ling.uni-duesseldorf.de)'),
writeln('%%%'),
writeln('%%% LOAD   GEN   GRAMMAR   WITH E.G. ?- genload hessian. <Return>'), 
writeln('%%% GENERATE   FROM  INPUT WITH E.G. ?- gen [h,''O'',n,d]. <Return>'),
writeln('%%% EXIT OT SIMPLE / GEN   WITH      ?- halt. <Return>'),
writeln('%%% THIS    HELP  MESSAGE  WITH      ?- help. <Return>'),
writeln('').


% --- support predicates

writeln(Term) :- nl, write(Term).

:- help.
\end{verbatim}
\paragraph{\Large GEN}
\begin{verbatim}
#!/bin/sh
# GEN Generate candidates from grammar, input and epenthesis count
# ===
# Usage: GEN "grammarfile, InputAsPrologList, MaxEpenthetics"
# First, issue compile statement for gen.pl, then compose three-parameter
# gen/3 goal (e.g. "hessian, [h,'O',n,d], 1" as instantiation for $1);
# send these directives to BinProlog (bp) with high quietness level (-q 6);
# remove superfluous Prolog prompt (?- ) and response (yes) as well as
# Prolog comment-initial (%%%) and stray empty lines (^$) with 'sed' afterwards

echo "compile(gen). gen($1)." | \
bp -q 6 | \
sed -e 's/?- //g;s/^yes//g;s/^%%%.*$//;/^$/d'
\end{verbatim}
\paragraph{\Large CON}
\begin{verbatim}
#!/bin/sh
# CON   implement constraint action
# ===
# Usage: CON Con1 Con2 ... ConN
# Con i denotes the name of a constraint script;
# We want to connect all of these ($*) as arguments to a single
# 'sed' invocation; since the syntax is
# sed -f scriptfile1 -f scriptfile2 ... -f scriptfileN,
# we use *another* invocation (`echo ... | sed ...`) of sed to 
# intersperse these  -f strings via substitution of the blanks
# argument separators; with the argument specification thus refined we
# can finally invoke 'sed' (the first one in the line) to act on the 
# constraint scripts

sed `echo " $*" | sed -e 's/ / -f /g'`
\end{verbatim}
\paragraph{\Large EVAL}
\begin{verbatim}
#!/bin/sh
# EVAL  evaluate CON-annotated output candidates
# ====
# Usage: stream_of_candidates | EVAL
# violation vector column is identifiable through field separator character '.'
# The column to sort on is the second column (first is the candidate itself)

sort -t. +1 -2
\end{verbatim}
\end{document}